\documentclass[sigconf]{acmart}
\usepackage{tabularx}
\usepackage{amsmath}
\usepackage{multirow}
\AtBeginDocument{%
  }

\setcopyright{acmlicensed}
\copyrightyear{2025}
\acmYear{2025}
\acmDOI{XXXXXXX.XXXXXXX}

\acmConference[KDD '25]{Make sure to enter the correct
  conference title from your rights confirmation emai}{August 03--07,
  2025}{Toronto, ON, Canada}
\acmISBN{978-1-4503-XXXX-X/18/06}

\begin{document}

\title{Diffusion-Inspired Cold Start with Sufficient Prior in Computerized Adaptive Testing}

\author{Haiping Ma}
\affiliation{%
  \institution{Information Materials and Intelligent Sensing Laboratory of Anhui Province, the Institutes of Physical Science and Information Technology, Anhui University}
  \city{Hefei}
  \state{Anhui}
  \country{China}
}
\email{hpma@ahu.edu.cn}

\author{Aoqing Xia}
\affiliation{%
  \institution{The Institutes of Physical Science and Information Technology, Anhui University}
  \city{Hefei}
  \state{Anhui}
  \country{China}
}
\email{q23301252@stu.ahu.edu.cn}

\author{Changqian Wang}
\affiliation{%
  \institution{The Institutes of Physical Science and Information Technology, Anhui University}
   \city{Hefei}
  \state{Anhui}
  \country{China}
}
\email{changqian.wang.dl@gmail.com}

\author{Hai Wang}
\affiliation{%
  \institution{The Institutes of Physical Science and Information Technology, Anhui University}
  \city{Hefei}
  \state{Anhui}
  \country{China}
}
\email{q22201135@stu.ahu.edu.cn}

\author{Xingyi Zhang}
\authornote{Corresponding author.}
\affiliation{%
  \institution{School of Computer Science and Technology, Anhui University}
  \city{Hefei}
  \state{Anhui}
  \country{China}}
\email{xyzhanghust@gmail.com}



\renewcommand{\shortauthors}{Haiping Ma et al.}

\begin{abstract}
Computerized Adaptive Testing (CAT) aims to select the most appropriate questions based on the examinee's ability and is widely used in online education. However, existing CAT systems often lack initial understanding of the examinee's ability, requiring random probing questions. This can lead to poorly matched questions, extending the test duration and negatively impacting the examinee's mindset, a phenomenon referred to as the Cold Start with Insufficient Prior~(CSIP) task. This issue occurs because CAT systems do not effectively utilize the abundant prior information about the examinee available from other courses on online platforms. These response records, due to the commonality of cognitive states across different knowledge domains, can provide valuable prior information for the target domain. However, no prior work has explored solutions for the CSIP task. In response to this gap, we propose \textbf{D}iffusion \textbf{C}ognitive \textbf{S}tates Transfe\textbf{R} Framework~(DCSR), a novel domain transfer framework based on Diffusion Models~(DMs) to address the CSIP task. Specifically, we construct a cognitive state transition bridge between domains, guided by the common cognitive states of examinees, encouraging the model to reconstruct the initial ability state in the target domain. To enrich the expressive power of the generated data, we analyze the causal relationships in the generation process from a causal perspective. Redundant and extraneous cognitive states can lead to limited transfer and negative transfer effects. Therefore, we designed three decoupling strategies to control confounding variables, thereby blocking backdoor paths that hinder causal discovery. Given that excessive uncertainty can affect the applicability of generated results to the CAT system, we propose consistency constraint and task-oriented constraint to control the randomness of the generated results and their relevance to the CAT task, respectively. Our DCSR can seamlessly apply the generated initial ability states in the target domain to existing question selection algorithms, thus improving the cold start performance of the CAT system. Extensive experiments conducted on five real-world datasets demonstrate that DCSR significantly outperforms existing baseline methods in addressing the CSIP task. The code is available at: \url{https://github.com/BIMK/Intelligent-Education/tree/main/DCSR}.
\end{abstract}

\ccsdesc[500]{Applied computing~Computer-assisted instruction}

\keywords{Computerized Adaptive Testing, Intellegent Education}


\maketitle
 \begin{figure}[!tbp]
 \centering
 \includegraphics[width=0.9\linewidth]{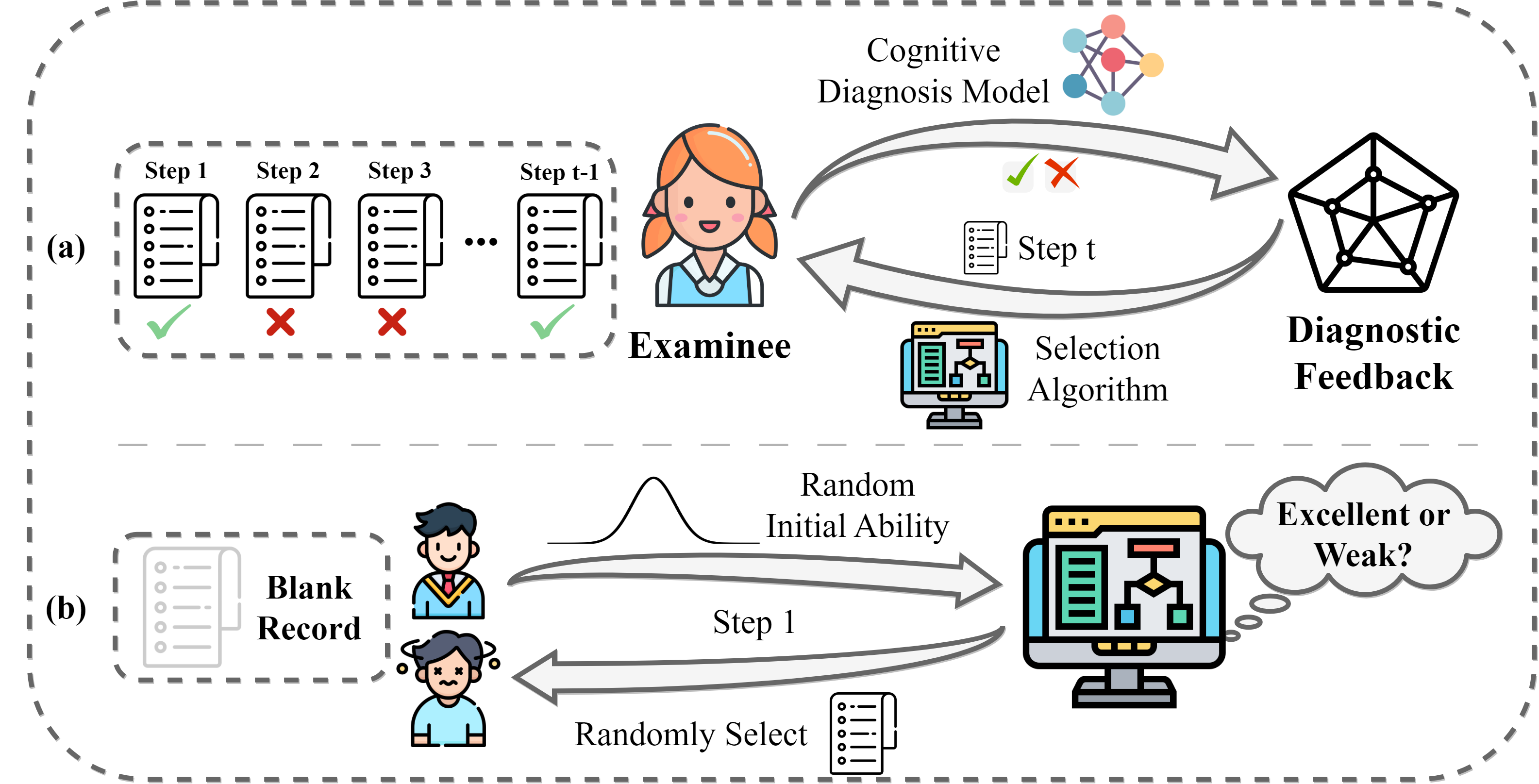}
\caption{Illustration of (a)~typical CAT process, and (b)~the dilemma of CAT under cold start.}
 \label{introduction}
\end{figure}
\section{Introduction}
As artificial intelligence empower education, computerized adaptive testing~(CAT) on online education platforms have garnered extensive attention~\cite{yang2024evolutionary, yang2024endowing, yu2024rdgt, yang2024evolutionary1}. CAT aims to provide examinees with a small number of appropriate questions to progressively assess their cognitive states in specific domains~\cite{intro1, intro2, intro3}. Typically, CAT consists of two iterative components: the cognitive diagnostic model~(CDM) and the question selection algorithm. As shown in Figure~\ref{introduction}~(a), the CDM estimates examinee's ability based on her response at step $t-1$~\cite{ma2024hd, yu2024rigl, yu2024disco, yang2023evolutionary}, after which the question selection algorithm provides suitable question for the next step, thus enabling a more accurate assessment of examinee's ability. Existing research on the question selection algorithm, a critical component of CAT, can be categorized into policy-based~\cite{maat, becat} and learnable-based~\cite{ncat, gmocat} approaches. The former generally converts the question selection process into model expectation, while the latter defines the CAT process as a bi-level optimization problem.

Despite the advancements made by various question selection algorithms, they still face challenges related to the cold start problem in the CAT process. As illustrated in Figure~\ref{introduction}~(b), on an online education platform, the system initially has no knowledge of the examinee's ability, causing the CAT system to tentatively select questions randomly to determine the starting point of the test.
The difficulty and content of the question might not align with the examinee's actual ability, leading to confusion or frustration and increasing the time cost of the assessment. Additionally, the question selection process is an iterative Markov chain. This means that if the question selected in the initial stages do not match the examinee's ability, it may lead to biased question selections in subsequent stages. Most question selection algorithms based on greedy strategies tend to amplify this bias, harming the performance of the CAT system. These algorithms, although capable of quickly finding local optima in certain scenarios, can easily get trapped in local optima over multiple rounds of question selection. 
In our paper, we refer to this task as Cold Start with Insufficient Priors~(CSIP), which arises from the insufficient understanding of the examinee by the question selection algorithm at the initial stage of the test, leading to uncertain starting points for  selection and subsequently affecting the adaptive testing process.

With the widespread application of educational platforms, vast amounts of examinee data are collected and stored~\cite{hu2023ptadisc}, providing valuable resources to enrich the prior information for CAT system. Notably, historical response records of examinees in other courses can serve as crucial references for CAT system to preliminarily understand examinee abilities. There are certain commonalities among knowledge concepts across different courses. For example, many physics questions, such as those in kinematics and dynamics, require algebraic equation solving, which is fundamental in mathematics. Therefore, an examinee's response records across multiple courses can reflect their foundational abilities and cognitive commonalities in multiple interdisciplinary fields, which are transferable attributes. These pre-diagnosed abilities can provide rich prior information for the target~(cold start)~course, allowing the CAT system to grasp the examinee's ability range in advance. For example, if an examinee performs well in mathematics, especially in complex algebra and geometry questions, the CAT system can infer that he also possess strong logical thinking and problem-solving skills in physics, thereby assigning a higher initial ability. To our knowledge, no research has yet utilized these cross-course data to solve the CSIP task, despite its significant and practical importance.

To fully leverage these cross-course response records and provide prior information about examinees to CAT system, we integrate and transfer the pre-diagnosis results of examinees to the target domain based on the concept of Diffusion Models~(DMs)~\cite{diffusion, diffusion1}. DMs add noise incrementally and then reconstruct the corrupted data step by step in reverse, which not only addresses noisy response records in the source domain but also aptly meets the needs of solving the CSIP task. This is due to the ability of DMs to gradually merge complex multi-distribution cross-domain data into a unified latent space during denoising, generating a unified latent representation, which reconstructs the initial ability of the target domain with specific transfer attributes from the noise. However, despite the great potential of using DMs to address the CSIP task, we uncover challenges in terms of what and how:~(1)~What kind of source domain information can be injected into the reverse denoising process as guidance.
~And (2)~How to constrain the output to match the CAT system, implying that excessive uncertainty in the generated results can lead to data unsuitable for CAT task.

To address these challenges, we design a novel domain transfer framework named the \textbf{D}iffusion \textbf{C}ognitive \textbf{S}tates Transfe\textbf{R} Framework~(DCSR) for the CSIP task. Guided by the principle of DMs, we generate initial abilities in the target domain for examinees to enhance the cold-start performance of CAT system. Specifically, to prevent the loss of personalized information in the diffusion generation results, we condition the denoiser on examinees' prior cognitive abilities to incorporate personalized target domain abilities. To avoid redundant information and negative transfer, we analyze the causal relationships between different variables from the causal perspective of model and design three decoupling strategies to adjust confounding variables in the backdoor path, including domain-shared cognition, domain-specific cognition, and orthogonal regularization-based gradient separation, blocking paths that obscure true causal relationships. To mitigate the excessive uncertainty of DMs and enhance the adaptability to CAT task, we design consistency constraint and task-oriented constraint to control randomness and match the input requirements of CAT system. The results generated by DCSR seamlessly integrate with existing question selection algorithms, improving the cold-start performance of CAT system. Extensive experimental results on five real-world datasets demonstrate that DCSR effectively addresses the CSIP task and significantly outperforms baselines.

\section{Related Work}
\subsection{Computerized Adaptive Testing}
Computerized Adaptive Testing~(CAT) aims to evaluate an examinee's ability progressively within a shorter test length~\cite{secat, dlcat}. CAT systems primarily consist of two components: \textbf{(1)~Cognitive Diagnostic Models~(CDMs)} for diagnosing examinees' abilities based on their responses to selected questions~\cite{ma2024dgcd, yang2024disengcd, cqw, liu2023homogeneous,yang2023cognitive}. Item Response Theory~(IRT)~\cite{irt}, a widely used CDM, assumes unidimensional independence and uses continuous latent variables to assess examinees' latent abilities.
With the widespread application of deep neural networks, neural CDMs such as NCD~\cite{ncd}, RCD~\cite{rcd}, and KaNCD~\cite{kancd} utilize neural networks to capture complex interactions between examinees and questions, enabling fine-grained diagnostic modeling. \textbf{(2)~Question selection algorithms} aim to adaptively choose the next appropriate question based on the CDM feedback. Early question selection algorithms were model-specific, such as IRT-specific Maximum Fisher Information~\cite{mfi}, Kullback-Leibler Information Index~\cite{kli}, and Max Entropy~\cite{cattheory}. However, these simple algorithms could not meet the evolving needs of CDMs, leading to performance limitations. Therefore, a model-agnostic question selection algorithm, MAAT~\cite{maat} was proposed, leveraging active learning to transform the question selection process into choosing questions with the greatest expected model change. Similarly, BECAT~\cite{becat} employs an expected gradient difference approach, treating the question selection process as a subset selection problem guided by theoretical estimates of examinees' true abilities. Another research direction focuses on data-driven question selection algorithms. For example, BOBCAT~\cite{bobcat}, NCAT~\cite{ncat}, and GMOCAT~\cite{gmocat} define the CAT task as a bi-level optimization problem, using reinforcement learning to learn question selection algorithm from large-scale response data. However, from a model perspective, the crucial CDM component in current CAT systems randomly initializes examinees' abilities at the test's outset. This random initialization necessitates more question selection steps for greedy algorithms to understand the examinee's ability range, thereby exacerbating the cold start problem in CAT system.

\subsection{Diffusion Model}
Diffusion Models~(DMs) have achieved impressive results in image generation~\cite{diffusion, diffusion1}. To transfer these successes to other domains, recent works~\cite{plugdiff, diffkg, denoisediff} have attempted to bridge the image domain with other fields. CODIGEM~\cite{codigem} was the first to extend the denoising module in DMs to recommendation systems. DiffuRec~\cite{diffurec}, DreamRec~\cite{dreamrec} and CF-Diff~\cite{cfdiff} model the latent representations of items and user preferences, guiding the denoising module to generate personalized representations. DiffRec~\cite{diffrec} and DiffuASR~\cite{diffuasr} not only reduce generation costs but also achieve temporal modeling of interaction sequences. Moreover, more research leverages the Markov chain modeling characteristics of DMs to explore sequence modeling, with few works utilizing diffusion features to study cross-domain problems. Although DiffCDR~\cite{diffcdr} has made preliminary explorations in this area, it introduces redundant information and faces limitations in cross-domain representation capacity due to the diverse entities in CAT system.
 \section{Preliminary}
\begin{figure*}[htp]
\centering
\includegraphics[width=1\textwidth]{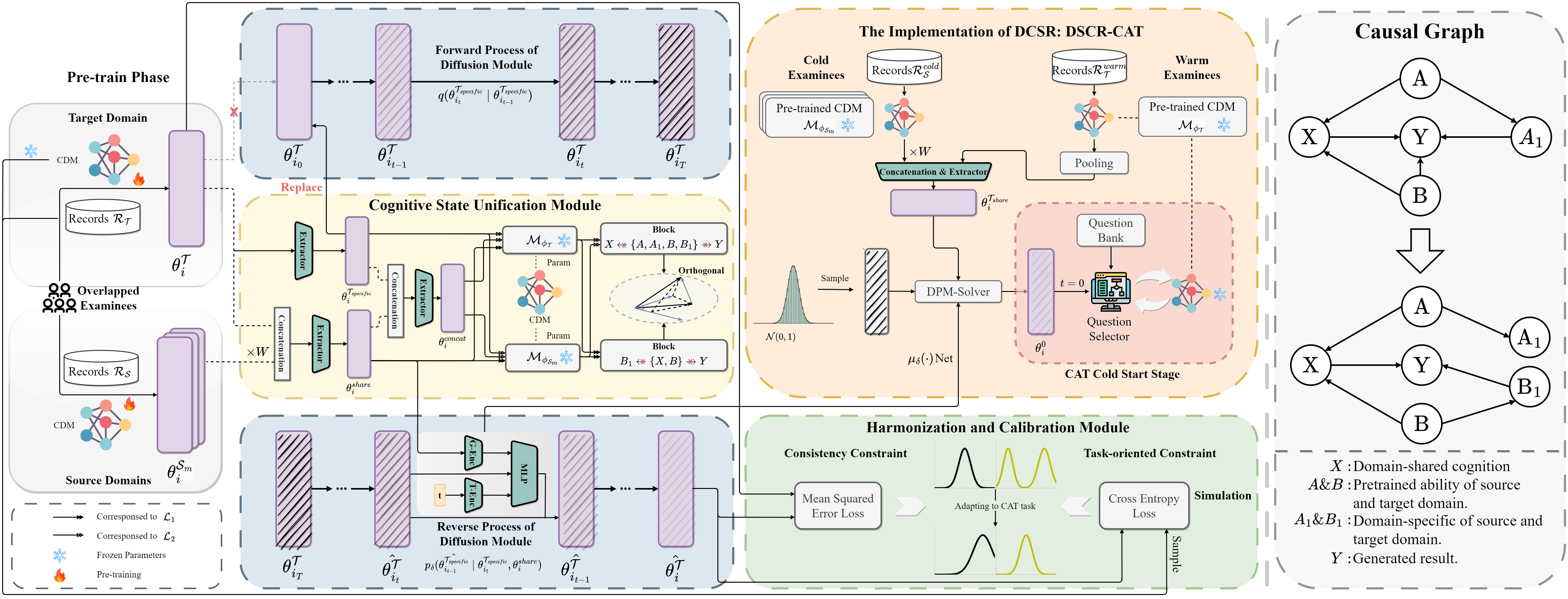}
\caption{The overview of DCSR: The left side corresponds to pre-training. While the blue, yellow, and green modules are used for training, and the orange modules correspond to the application in CAT. Additionally, the right side depicts the causal discovery in the generation process.}
\label{framework}
\end{figure*}
\subsection{Computerized Adaptive Testing}
\subsubsection{Task Introduction.}
In an online education platform, the Computerized Adaptive Testing~(CAT) system adaptively selects questions for examinees to accurately reveal their cognitive abilities. The CAT system comprises two components:~(1)~a Cognitive Diagnosis Model~(CDM)~$\mathcal{M} $, which assesses the examinee's ability state after responding to the selected questions, and~(2)~a question selection algorithm $\Pi$, which chooses the next most informative question based on the examinee's current ability state to measure their ability more precisely. These two components alternately iterate until a termination condition is met. 

Specifically, given a examinee $e_i$ and his/her candidate question set $\mathcal{Q}_i$, the CAT process can be represented as:
\begin{equation}
 \begin{aligned}
 &q^t\leftarrow\Pi(Q_i\mid \theta_i^{t-1}),\\
 &\theta_i^t \leftarrow \mathcal{M}(r_{iq^t}\mid q^t,\theta_{i}^{t-1}),
 \end{aligned}
\end{equation}
where $q^t$ denotes the question selected by $\Pi$ based on the examinee's ability $\theta_i^{t-1}$ at time $t-1$ from the candidate question set $\mathcal{Q}_i$, which only includes questions that the examinee $e_i$ has not yet responded to. The variable $r_{iq^t}$ represents the response result of examinee $e_i$ to question $q^t$. Note that $T$ is the maximum number of steps for terminating the test. 

\subsubsection{Cold start with insufficient priors.}
In the Cold Start with Insufficient Priors~(CSIP) scenario, the question selection algorithm considers only the information from the target domain while ignoring the rich data from the source domains. To alleviate this lack of prior information, we define $M$ source domains $\mathcal{S}_1,\mathcal{S}_2,\dots,\mathcal{S}_M$ and one target domain $\mathcal{T}$ in the Diffusion Cognitive State Transfer Framework~(DCSR). Each domain includes three groups of entities. We use $\mathcal{E}_{\mathcal{S}_m}$, $\mathcal{Q}_{\mathcal{S}_m}$ and $\mathcal{C}_{\mathcal{S}_m}$ to denote the sets of examinees, questions, and knowledge concepts in the $m\in M$ source domain, respectively. Similarly, $\mathcal{E}_\mathcal{T}$, $\mathcal{Q}_\mathcal{T}$ and $\mathcal{C}_\mathcal{T}$ denote the three entities in the target domain. The overlapping examinees between domains are defined as $\mathcal{E}_{\mathcal{O}}\ni \mathcal{E}_{\mathcal{O}_m}=\mathcal{E}_{\mathcal{S}_m}\cap \mathcal{E}_\mathcal{T}$, while the other two entity groups generally do not have overlapping elements across domains. 

\subsubsection{Training and Testing Phases.}
\sloppy In the given CAT testing platform, both warm-start and cold-start, meaning the CAT system has never encountered before, are included in the target domain. Their response records can be represented as $\smash{\mathcal{R}_\mathcal{T}=\left\{ \mathcal{R}^{warm}_\mathcal{T},\mathcal{R}^{cold}_\mathcal{T} \right\}=\left\{ ( e_i,q_j,r_{ij} ) \mid e_i \in \mathcal{E}_\mathcal{T}^{warm}\cup\mathcal{E}_\mathcal{T}^{cold} \right\}}$, where $r_{ij} = 1$ indicates that examinee $e_i$ answered question $q_j\in \mathcal{Q}_\mathcal{T}$ correctly, and $r_{ij} = 0$ otherwise. Similarly, the response records in the $m$ source domain can be represented as $\smash{\mathcal{R}_{\mathcal{S}_m}=\left\{(e_i,q_j,r_{ij})\mid e_i\in\mathcal{E}_{\mathcal{O}_m},q_j\in\mathcal{Q}_{\mathcal{Q}_m}\right\}}$, and all source domain records are denoted as $\smash{\mathcal{R}_\mathcal{S}=\left\{\mathcal{R}_{\mathcal{S}_1},\mathcal{R}_{\mathcal{S}_2},\dots,\mathcal{R}_{\mathcal{S}_m}\right\}}$. To prevent data leakage, the aforementioned response sets are uniformly divided into a training set $\smash{\mathcal{D}_{train}=\{\mathcal{R}_\mathcal{T}^{warm},\mathcal{R}_\mathcal{S}\}}$ and a test set $\smash{\mathcal{D}_{test}=\mathcal{R}_\mathcal{T}^{cold}}$. 

Our DCSR consists of two phases: training the initial abilities of examinees in the source domain based on $\mathcal{D}_{train}$ and testing the performance of question selection in the CAT system without learning, based on $\mathcal{D}_{test}$. 

\subsubsection{Pre-Establish Cognitive States}
The data in the training set $\mathcal{D}_{train}$ comes from the response records of overlapping examinees $\mathcal{E}_\mathcal{O}$ between domains. We pre-train these examinees to establish their cognitive states in both the source and target domains. The pre-training process is expressed as: 
\begin{equation}
 \theta^*= \mathop{\arg\min}\limits_{\theta \in \Theta}
\sum_{(e_i,q_j,r_{ij}) \in \mathcal{D}_{train} }\mathcal{L}\left (r_{ij}, \mathcal{M}_\Psi(q_j|\theta_i) \right ),
\end{equation}
\sloppy where $\mathcal{M}_\Psi$ is the CDM used for pre-training, and different domain data correspond to different CDM parameters, i.e., $\Psi=\{\psi_{\mathcal{S}_1},\psi_{\mathcal{S}_2}, \dots,\psi_{\mathcal{S}_M},\psi_{\mathcal{T}}\}$. The abilities of overlapping examinees are denoted as $\Theta_{\mathcal{S}}=\left\{\theta_i\mid i\in \mathcal{E}_\mathcal{O}\right\}$, and similarly, the abilities in the target domain are denoted as $\Theta_{\mathcal{T}}$. 

\subsection{Diffusion Model}
Diffusion Models~(DMs) have demonstrated exceptional performance in fields such as computer vision~\cite{diffusion}. Typically, DMs consist of two parts: the forward process and the reverse process.

\subsubsection{Forward Process.}
Given a data point $x_0\sim q(x)$ sampled from the true data distribution, the forward process gradually degrades $x_0$ into standard Gaussian noise $x_T \sim \mathcal{N}(0,1)$ by injecting Gaussian noise over $T$ steps. Specifically, the process of converting $x_{t-1}$ to $x_t$ in DMs is represented as $q(x_t \mid x_{t-1}) = \mathcal{N}(x_t; \sqrt{1-\beta_t} x_{t-1}, \beta_t \mathrm{I})$, where $t\in\left\{1,\dots,T\right\}$ represents the diffusion steps, $\beta_t \in (0, 1)$ is the predefined noise scheduling coefficient, and $\mathcal{N}$ denotes the Gaussian distribution.

\subsubsection{Reverse Process.} \label{sec:prereverse}
In the reverse process, DMs learn to remove the added noise, thereby recovering the original data distribution $x_0$ from pure noise, aiming to introduce minor uncertainties in the generation process. This process learns a parameterized network $p_\delta (x_{t-1}\mid x_t)$ to approximate the reverse process, which can be formalized as $p_\delta (x_{t-1}\mid x_t)=\mathcal{N}(x_{t-1};\mu_\delta (x_t,t),\Sigma_\delta(x_t,t))$, where $\mu_\delta$ and $\Sigma_\delta$ are the mean and variance of the Gaussian distribution predicted by a neural network with parameters $\delta$.

\section{Method}

\textbf{Overview. }
The core idea of this work is to transfer prior diagnostic results from the source domain to the target domain, thereby generating personalized initial abilities for cold-start examinees in the target domain course. As illustrated in Figure~\ref{framework}, our DCSR is built upon the Diffusion Module and is supported by two key components: the Cognitive State Unification Module~(CSUM) and the Harmonization and Calibration Module~(HCM). Specifically, the Diffusion Module uses prior abilities from the source domain to reconstruct the cognitive state of examinees in the target domain from noise. From a causal model perspective, the generated abilities are influenced by redundant knowledge, and domain-specific cognition may cause negative transfer. Therefore, the CSUM is employed to control confounding variables in the backdoor path, thereby uncovering the true causal relationships. Additionally, to mitigate uncertainty during the diffusion process, we designed consistency and task-oriented constraints in HCM, aiming to ensure that the generated results adhere to the true distribution and meet the requirements of CAT task. It is worth noting that our framework demonstrates significant scalability, seamlessly integrating the generated initial abilities in the target domain into existing CAT, thereby improving its cold-start performance. 
\vspace{-2mm}
\subsection{Diffusion Module} \label{sec:diffusionmodule}
We employ the concept of diffusion as the backbone of our model, establishing a bridge for cognitive state transformation between the source and target domains. The purpose of the Diffusion Module is to incorporate prior information from the source domain into the initial ability estimation process in the target domain. Therefore, it inherently involves two distinct processes: the forward noise addition process and the reverse denoising process.

During training, we gradually inject Gaussian noise into the ability vectors $\Theta_\mathcal{T}$ of examinees in the target domain over $T$ steps. For a specific examinee $e_i\in\mathcal{E}_{\mathcal{O}}$, his ability vector $\theta_{i_0}^\mathcal{T}=\theta_{i}^\mathcal{T}\in\Theta_\mathcal{T}$ is corrupted into $\theta_{i_{1:T}}^\mathcal{T}$, which is modeled as a Gaussian transition Markov chain: 
\begin{equation}
 \setlength{\abovedisplayskip}{3pt}
\setlength{\belowdisplayskip}{3pt}
q(\theta_{i_{1:T}}^\mathcal{T}\mid\theta_{i_0}^\mathcal{T})=\prod_{t=1}^{T}\mathcal{N}(\theta_{i_t}^\mathcal{T};\sqrt{1-\beta_t}\theta_{i_{t-1}}^\mathcal{T},\beta_t\mathrm{I})
\end{equation}
where $\beta_t\in(0,1)$ controls the scale of noise added at the \$t\$-th step, and $\mathcal{N}$ denotes a Gaussian distribution.

In the reverse denoising process, the traditional denoising methods~(as introduced in section~\ref{sec:prereverse}) are inadequate for effective transfer because the denoising process modeled lacks guidance from source domain information, resulting in the loss of personalization in the generated abilities. To generate personalized ability vectors, we propose utilizing prior information from the source domain to guide the denoising process. Specifically, for examinee $e_i$, we use the pretrained source domain ability $\theta_i^{\mathcal{S}}\in\Theta_{\mathcal{S}}$ as guidance: 
\begin{equation}
 \setlength{\abovedisplayskip}{5pt}
\setlength{\belowdisplayskip}{5pt}
 p_\delta(\hat{\theta_{i_{t-1}}^\mathcal{T}}\mid \theta_{i_t}^\mathcal{T},\theta_i^{\mathcal{S}})= \mathcal{N}(\hat{\theta_{i_{t-1}}^\mathcal{T}};\mu_\delta(\theta_{i_t}^\mathcal{T},\theta_i^{\mathcal{S}},t),\Sigma_\delta(\theta_{i_t}^\mathcal{T},\theta_i^{\mathcal{S}},t)),
\end{equation}
where $\mu_\delta$ and $\Sigma_\delta$ are the parameters output by neural networks with learnable parameters $\delta$. 

\subsection{Cognitive State Unification Module}\label{sec:CSUM}
Although using unified source domain abilities obtained through CDM to guide the reverse denoising process is a promising approach, from the causal perspective of the model, as shown in the right side of Figure~\ref{framework}, the generated ability $Y$ is influenced by domain-shared cognition $X$, domain-specific cognition $A_1$, and target domain ability $B$ during the training phase. This is because we use random sampling of time steps for training, which retains some personalized information in the target domain ability vector even while introducing noise. In other words, $\theta_{i_{1:T}}^\mathcal{T}$ does not approximate standard Gaussian noise, which lacks extensive personalized features. However, the inclusion of domain-specific cognition $A_1$ increases the complexity of the model and reduces its generalization ability, making the model prone to overfitting. Therefore, we propose to explore the causal relationship between the generated result $Y$ and the domain-shared cognition $X$ and the pretrained ability $B$ of the target domain, which can also be further decoupled into domain-specific cognition $B_1$ and domain-shared cognition $X$. Next, we will specifically analyze the impact of different factors on the generated result. 

First, there are three paths between domain-shared cognition $X$ and the generated results $Y$: $X\to Y$, $X \gets \{A , A_1\}\to Y$, and $X \gets \{B , B_1\}\to Y$, where the latter two paths are confounding paths. We will apply the backdoor criterion to explore the causal relationship of $X\to Y$, which means we need to control the confounding variables $C=\{A,A_1\}\cup\{B,B_1\}$ to block the backdoor paths: 
\begin{equation}
 \setlength{\abovedisplayskip}{5pt}
\setlength{\belowdisplayskip}{5pt}
 P(Y\mid do(X))=\sum_{C}P(Y\mid X,C=c)P(C=c),
\end{equation}
where the first term represents the effect of $X$ on $Y$ while controlling for the confounding variables $C$, and the second term represents the joint probability distribution of the confounding variables. To control the confounding variables, inspired by~\cite{0123}, we decouple the domain-shared cognition $\theta_i^{share}$ from all the prior abilities in the source and target domains: 
\begin{equation}
 \setlength{\abovedisplayskip}{5pt}
\setlength{\belowdisplayskip}{5pt}
    \theta_i^{share}= f_{\varphi _2}(\sigma(f_{\varphi _1}(\theta_i^\mathcal{T}\parallel 
\mathbb{E}_{m\sim \mathcal{S}_M}\mathrm{W}_{m}\theta_i^{\mathcal{S}_m}))),\label{eq:specific}
\end{equation}
where $W\in \mathbb{R}^{ M \times d}$ is a weight matrix used to map the pretrained abilities from multiple domains into the same feature space, where $d$ is the feature dimension related to CDM, $ f_{\varphi _1}$ and $ f_{\varphi _2}$ are linear layers with different parameters, $\sigma$ refers to the activation function, and $(\cdot\parallel\cdot)$ denotes the concat operation. We then encourage this decoupled domain-shared cognition to assist in predicting responses across all domains: 
\begin{equation}
\setlength{\abovedisplayskip}{5pt}
\setlength{\belowdisplayskip}{5pt}
 \begin{aligned}
 \mathcal{L}_1 =& \sum_{(e_i,q_j^\mathcal{T},r_{ij}^\mathcal{T})\in\mathcal{R}^{warm}_\mathcal{T}}\| r_{ij}^\mathcal{T} -\mathcal{M}_{\phi_T}(\theta_i^{share},q_j^\mathcal{T}) \|\\[0.0mm]
&+\sum_{\mathcal{S}_{m=1}}^M\sum_{(e_i,q_j^{\mathcal{S}_m},r_{ij}^{\mathcal{S}_m})\in\mathcal{R}_{\mathcal{S}_m}}\| r_{ik}^{\mathcal{S}_m} -\mathcal{M}_{\phi_{S_m}}(\theta_i^{share},q_k^{\mathcal{S}_m}) \|,
 \end{aligned}
\end{equation}
where $\theta^{share}$ is encouraged to predict responses across all scenarios, including the target domain in the first term and all source domain response data in the second term. 
Next, the specific cognition of the target domain $B_1$ has two paths influencing the generated results: $B_1\to Y$ and $B_1\gets\{X,B\}\to Y$. We adjust the confounding variable $\{X,B\}$ in the second confounding path to block the backdoor path. Specifically, we extract the specific cognition of the target domain, which will be used as input in the section~\ref{sec:diffusionmodule} alongside obtaining the overall cognitive state of the target domain~(corresponding to event $B$ in the causal graph): 
\begin{equation}
\setlength{\abovedisplayskip}{5pt}
\setlength{\belowdisplayskip}{5pt}
\begin{aligned}
 &\theta_i^{\mathcal{T}_{specific}}= f_{\varphi _4}(\sigma(f_{\varphi _3}(\theta^\mathcal{T}))),\\[0.0mm]
 &\theta_i^{concat}=f_{\varphi _5}(\sigma(f_{\varphi _6}(\theta_i^{\mathcal{T}_{specific}}\parallel \theta_i^{share}))),
\end{aligned}
\end{equation}
where $ f_{\{\varphi _3,\varphi _4,\varphi _5,\varphi _6\}}$ are four linear layers with different parameters. To learn the specific cognition of the target domain, we propose a novel decoupling strategy, which encourages the prediction of within-domain response results by domain-specific cognition while degrading performance in other domains: 
\begin{equation}
\setlength{\abovedisplayskip}{5pt}
\setlength{\belowdisplayskip}{5pt}
 \begin{aligned}
 \mathcal{L}_2 &= \sum_{(e_i,q_j^\mathcal{T},r_{ij}^\mathcal{T})\in\mathcal{R}^{warm}_\mathcal{T}}\Big(\| r_{ij}^\mathcal{T} -\mathcal{M}_{\phi_T}(\theta_i^{concat},q_j^\mathcal{T}) \|\\[0.0mm]
 &\quad\quad\quad\quad\quad\quad\quad\quad+\| r_{ij}^\mathcal{T} -\mathcal{M}_{\phi_T}(\theta_i^{\mathcal{T}_{specific}},q_j^\mathcal{T}) \|\Big)\\[0.0mm]
 &-\sum_{\mathcal{S}_{m=1}}^M\sum_{(e_i,q_j^{\mathcal{S}_m},r_{ij}^{\mathcal{S}_m})\in\mathcal{R}_{\mathcal{S}_m}}\Big(\| r_{ik}^{\mathcal{S}_m} -\mathcal{M}_{\phi_{S_m}}(\theta_i^{concat},q_k^{\mathcal{S}_m}) \| \\[0.0mm]
 &\quad\quad\quad\quad\quad\quad\quad\quad\quad\quad+\| r_{ik}^{\mathcal{S}_m} -\mathcal{M}_{\phi_{S_m}}(\theta_i^{\mathcal{T}_{specific}},q_k^{\mathcal{S}_m}) \|\Big)\\[0.0mm]
 &+\mathbb{E}_{e_i\in\mathcal{E}_{\mathcal{O}_m}}\|\theta_i^\mathcal{T}-\theta_i^{concat}\|.
 \end{aligned}
\end{equation}
Here, the first term encourages the target domain-specific cognition to assist in predicting the response records in target domain, both independently and in conjunction with domain-shared cognition. The second term aims to intentionally degrade prediction performance in the source domains, indicating that the target domain-specific cognition is not applicable to the source domains. The third term constrains the diagnostic abilities obtained from pre-training. Additionally, to further control the influence of confounding variables on the generated results, we apply gradient-based orthogonal regularization~\cite{orth} to ensure the independence of the above representations in the feature space:
\begin{equation}
\setlength{\abovedisplayskip}{5pt}
\setlength{\belowdisplayskip}{5pt}
 \mathcal{L}_3 = \Big\|\frac{\nabla _{\varphi _1,\varphi _2}\mathcal{L}_1}{\|\nabla_{\varphi _1,\varphi _2} \mathcal{L}_1\|}\cdot \frac{\nabla _{\varphi _3,\varphi _4}\mathcal{L}_2}{\|\nabla_{\varphi _3,\varphi _4} \mathcal{L}_2\|}\Big\|_2.
\end{equation}
This approach minimizes the influence of domain-specific cognition when learning domain-shared cognition, and vice versa.

Therefore, the parameters in the forward and reverse processes are updated to $\smash{q(\theta_{i_t}^{\mathcal{T}_{specific}}\mid \theta_{i_{t-1}}^{\mathcal{T}_{specific}})}$ and $\smash{p_\delta(\hat{\theta_{i_{t-1}}^{\mathcal{T}_{specific}}}\mid\theta_{i_{t}}^{\mathcal{T}_{specific}},\theta_i^{share})}$, respectively.

\subsection{Harmonization and Calibration Module}\label{sec:HCM}
To learn the parameters $\delta$ of the denoising network, DCSR aims to maximize the Evidence Lower Bound~(ELBO) of the observed examinee ability $\theta^{\mathcal{T}_{specifc}}_i$:
\begin{equation}
 \setlength{\abovedisplayskip}{3pt}
\setlength{\belowdisplayskip}{3pt}
\resizebox{0.95\hsize}{!}{$
 \begin{aligned}
 &\mathrm{log}~p(\theta^{\mathcal{T}_{specific}}_i)\ge \underbrace{\mathbb{E}_{q(\theta^{\mathcal{T}_{specific}}_{i_0}\mid \theta^{\mathcal{T}_{specific}}_{i_1})}\left [ \mathrm{log}~p_\delta(\theta^{\mathcal{T}_{specific}}_{i_0}\mid \theta^{\mathcal{T}_{specific}}_{i_1}) \right ]}_{\textbf{:=}
\mathcal{L}_0} \\ 
 & -\sum_{t=2}^T\underbrace{\mathbb{E}_{q(\theta^{\mathcal{T}_{specific}}_{i_t}\mid \theta^{\mathcal{T}_{specific}}_{i_0})}\left[\mathrm{KL(q(\theta^{\mathcal{T}_{specific}}_{i_{t-1}}\mid\theta^{\mathcal{T}_{specific}}_{i_t},\theta^{\mathcal{T}_{specific}}_{i_0}))\parallel p_\delta(\theta^{\mathcal{T}_{specific}}_{i_{t-1}}\mid \theta^{\mathcal{T}_{specific}}_{i_t})}\right]}_{\textbf{:=}
\mathcal{L}_{t-1}}.
 \end{aligned}
 $}
\end{equation}
Here, the first term represents the reconstruction term, which recovers the probability of $\theta^{\mathcal{T}_{specific}}_{i_0}$. The second term is the denoising matching term, which aligns the intractable posterior probability $p_\delta(\cdot)$ with the tractable distribution $q(\cdot)$. To maintain training stability and simplify computation, we ignore the learning of $\Sigma_\delta(\cdot)$ and set it to a fixed value$\beta_t$ as in the forward process~\cite{diffusion, diffrec}. The denoising matching term can then be further computed as:
\begin{equation}
 \setlength{\abovedisplayskip}{3pt}
\setlength{\belowdisplayskip}{3pt}
\begin{aligned}
 \mathcal{L}_{t-1} = \mathbb{E}_{q(\theta^{\mathcal{T}_{specific}}_{i_0}\mid \theta^{\mathcal{T}_{specific}}_{i_1})}&\left[ \frac{1}{2\beta_t} \left\|\mu_\delta(\theta^{\mathcal{T}_{specific}}_{i_t},\theta^{share},t) \right. \right. \\
 &\left. \left. - \widetilde{\mu}(\theta^{\mathcal{T}_{specific}}_{i_t},\theta^{\mathcal{T}_{specific}}_{i_0}) \right\| \right],
\end{aligned}
\end{equation}
where $\widetilde{\mu}(\cdot) $ satisfies the tractable distribution $\smash{q(\theta^{\mathcal{T}_{specific}}_{i_{t-1}}\mid\theta^{\mathcal{T}_{specific}}_{i_t},\theta^{\mathcal{T}_{specific}}_{i_0})=\mathcal{N}(\theta^{\mathcal{T}_{specific}}_{i_{t-1}};\widetilde{\mu}(\cdot),\beta_t\mathrm{I})}$.

Although the above training objective ensures diversity in generated samples, excessive randomness introduced by the diffusion model during training and inference stages may result in generated outcomes that do not match the requirements of the CAT system. Therefore, we constrain the generated ability vectors from two aspects: Consistency Constraint and Task-oriented Constraint.

\subsubsection{Consistency Constraint} aims to minimize the difference between generated vectors and real data samples in the feature space, thus limiting uncertainty within a controllable range. Specifically, the Diffusion Module diffuses the domain-shared cognitive features into domain-specific cognitive features. Firstly, the general cognitive features of the target domain are generated as $\hat{\theta_i^\mathcal{T}}$:
\begin{equation}
 \setlength{\abovedisplayskip}{3pt}
\setlength{\belowdisplayskip}{3pt}
\hat{\theta_i^\mathcal{T}}=\mu_\delta(\theta_{i_{T:1}}^{\mathcal{T}_{specific}},\theta_i^{share},t_{T:1})+\beta_t\epsilon,\epsilon\in(0,1).
\end{equation}
Due to inherent randomness, the generated ability vectors are noisy, which is not conducive to reflecting the examinee's true ability. Thus, the consistency constraint aims to minimize the difference between $\hat{\theta_i^\mathcal{T}}$ and the target domain ability features $\theta_i^\mathcal{T}$ obtained through pre-training, which can be modeled as:
\begin{equation}
  \setlength{\abovedisplayskip}{3pt}
\setlength{\belowdisplayskip}{3pt}\mathcal{L}_{cc}=\mathbb{E}_{e_i\in\mathcal{E}_{\mathcal{O}}}(\theta_i^\mathcal{T}-\hat{\theta_i^\mathcal{T}})^2.
\end{equation}

 \subsubsection{Task-oriented Constraint} ensures that the generated ability features meet the requirements of the CAT task. Specifically, we sample some questions $q_j\in\mathcal{R}_\mathcal{T}^{warm}$ from the training set of the target domain and match the generated target domain general cognitive features $\hat{\theta_i^\mathcal{T}}$ with the CDM:
\begin{equation}
 \setlength{\abovedisplayskip}{3pt}
\setlength{\belowdisplayskip}{3pt}
\begin{aligned}
 \nabla _{\hat{\theta^\mathcal{T}}}\mathcal{L}_{tc}=&-\sum_{(e_i,q_j,r_{ij})\in\mathcal{R}_\mathcal{T}^{warm}}\Big(r_{ij}\mathrm{log}\big(\mathcal{M}_{\psi_\mathcal{T}}(\hat{\theta_i^\mathcal{T}},q_j)\big)\\
 &+(1-r_{ij})\mathrm{log}\big(1-\mathcal{M}_{\psi_\mathcal{T}}(\hat{\theta_i^\mathcal{T}},q_j)\big)\Big),
\end{aligned}
\end{equation}
where $\mathcal{M}_{\psi_\mathcal{T}}(\cdot)$ denotes the pre-trained CDM for the target domain, with its parameters frozen, and only the network parameters $\mu_\delta(\cdot)$ of the denoising module are updated.

\subsection{DCSR-CAT: Implementation of DCSR}
In this section, we introduce DCSR-CAT as an implementation of DCSR to demonstrate its applicability to CAT.

During this stage, DCSR no longer involves the forward process but directly takes pure noise $\epsilon_0\sim \mathcal{N}(0,1)$ as input. For a given cold-start examinee $e_i\in\mathcal{R}_\mathcal{T}^{cold}$ who has response records only in the source domain, we use the pre-trained CDM to obtain their prior ability and calculate their domain-shared cognitive features $\theta_i^{share}$. It is noteworthy that since the original $\theta_i^\mathcal{T}$ is unknown, the average cognitive state of the target domain is used as a substitute:
\begin{equation}
  \setlength{\abovedisplayskip}{3pt}
\setlength{\belowdisplayskip}{3pt}\theta_i^{\mathcal{T}_{cold}}=\mathbb{E}_{e_j\in\mathcal{R}^{warm}_\mathcal{T}}\mathcal{M}_{{\psi_\mathcal{T}}_{\nabla_\theta} }(e_j),
\end{equation}
which retains domain information to some extent. Therefore, substituting back into equation~\eqref{eq:specific}, we obtain the shared cognitive ability $\theta_i^{share}$ of the cold-start examinee $e_i$. To improve inference efficiency, we use DPM-Solver~\cite{dpmsolver} as a fast solver to efficiently obtain the initial ability of cold-start examinee:
\begin{equation}
 \setlength{\abovedisplayskip}{3pt}
\setlength{\belowdisplayskip}{3pt}
 \theta_i^0=\mathrm{Solver}(\mu_\delta(\cdot),\theta_i^{share},\epsilon_0),
\end{equation}
which is crucial for real-world applications. Our DCSR seamlessly integrates with any existing question selection algorithm $\Pi$. At the beginning of the test, the selection process can be represented as:
\begin{equation}
 \setlength{\abovedisplayskip}{3pt}
\setlength{\belowdisplayskip}{3pt}
 q^1\leftarrow\Pi(Q_i\mid \theta_i^{0}),
\end{equation}
which indicates that the item selection algorithm $\Pi$ selects the appropriate item $q^1$ in step $t=1$ from the candidate item pool $\mathcal{Q}_i$ for examinee $e_i$ based on the initial ability $\theta_i^0$ initialized by DCSR. The subsequent steps follow the general CAT system workflow.
\begin{table}[!t]
\centering
  \renewcommand{\arraystretch}{1.}
 \setlength{\tabcolsep}{0.2mm}{
 \small
\begin{tabularx}{1\columnwidth}{@{} X X X X X X @{}}
\toprule
\textbf{Statistics} & \textbf{C}& \textbf{C++} & \textbf{DS}& \textbf{Java}& \textbf{Python}\\
\midrule
\#Examinee & 3,856& 3,856& 3,856& 3,856& 3,856\\
\#Question & 115,335& 10,241& 14,207& 10,043& 11,600\\
\#Concept & 654& 472& 560& 541& 480\\
\#Log & 2,163,624& 1,332,005& 1,398,444& 1,086,906& 923,879\\
\bottomrule
\end{tabularx}}
\caption{The statistics of the datasets.}
\label{dataset_table}
 \vspace{-10pt}
\end{table}
\vspace{-2mm}
\begin{table*}[ht]
\centering
  \renewcommand{\arraystretch}{1.5}
 \resizebox{\textwidth}{!}{%
\begin{tabular}{cc|cccccccccccccccccccc}
\hline
\toprule
    \toprule
\multicolumn{2}{c|}{CDM}                        & \multicolumn{8}{c|}{IRT}                                                                                                                                                                                                                                              & \multicolumn{12}{c}{NCD}                                                                                                                                                                                                                                                                                                                                                            \\ \hline
\multicolumn{2}{c|}{CAT}                        & \multicolumn{4}{c|}{Fisher}                                                                                                       & \multicolumn{4}{c|}{MAAT}                                                                                                         & \multicolumn{4}{c|}{MAAT}                                                                                                         & \multicolumn{4}{c|}{BECAT}                                                                                                        & \multicolumn{4}{c}{NCAT}                                                                                    \\ \hline
\multicolumn{2}{c|}{Metrics}                    & \multicolumn{20}{c}{AUC(\%)@1/ACC(\%)↑@1}                                                                                                                                                                                                                                                                                                                                                                                                                                                                                                                                                                                                                  \\ \hline
\multicolumn{2}{c|}{Baselines}                  & \multicolumn{1}{c|}{Random}    & \multicolumn{1}{c|}{MLCCM}     & \multicolumn{1}{c|}{DCSR}      & \multicolumn{1}{c|}{Oracle*}    & \multicolumn{1}{c|}{Random}    & \multicolumn{1}{c|}{MLCCM}     & \multicolumn{1}{c|}{DCSR}      & \multicolumn{1}{c|}{Oracle*}    & \multicolumn{1}{c|}{Random}    & \multicolumn{1}{c|}{MLCCM}     & \multicolumn{1}{c|}{DCSR}      & \multicolumn{1}{c|}{Oracle*}    & \multicolumn{1}{c|}{Random}    & \multicolumn{1}{c|}{MLCCM}     & \multicolumn{1}{c|}{DCSR}      & \multicolumn{1}{c|}{Oracle*}    & \multicolumn{1}{c|}{Random}    & \multicolumn{1}{c|}{MLCCM}     & \multicolumn{1}{c|}{DCSR}      & Oracle*    \\ \hline
\multicolumn{1}{c}{\multirow{2}{*}{C}}   & C++ & \multicolumn{1}{c|}{71.6/75.0} & \multicolumn{1}{c|}{78.5/77.5} & \multicolumn{1}{c|}{\textbf{77.9}/\textbf{76.3}} & \multicolumn{1}{c|}{79.5/75.4} & \multicolumn{1}{c|}{71.2/74.9} & \multicolumn{1}{c|}{\textbf{78.5}/\textbf{77.4}} & \multicolumn{1}{c|}{77.7/76.0} & \multicolumn{1}{c|}{79.6/75.7} & \multicolumn{1}{c|}{67.4/62.8} & \multicolumn{1}{c|}{74.7/74.6} & \multicolumn{1}{c|}{\textbf{79.2}/\textbf{77.3}} & \multicolumn{1}{c|}{88.0/83.1} & \multicolumn{1}{c|}{67.4/62.7} & \multicolumn{1}{c|}{74.6/74.6} & \multicolumn{1}{c|}{\textbf{79.2}/\textbf{77.3}} & \multicolumn{1}{c|}{88.0/83.0} & \multicolumn{1}{c|}{67.4/62.8} & \multicolumn{1}{c|}{74.5/74.6} & \multicolumn{1}{c|}{\textbf{79.3}/\textbf{77.3}} & 88.0/83.1 \\ \cline{2-22} 
\multicolumn{1}{c}{}                     & DS  & \multicolumn{1}{c|}{71.8/75.4} & \multicolumn{1}{c|}{\textbf{75.5}/\textbf{77.0}} & \multicolumn{1}{c|}{75.4/75.9} & \multicolumn{1}{c|}{77.2/78.2} & \multicolumn{1}{c|}{71.1/75.2} & \multicolumn{1}{c|}{74.9/\textbf{76.8}} & \multicolumn{1}{c|}{\textbf{75.0}/75.7} & \multicolumn{1}{c|}{77.2/78.2} & \multicolumn{1}{c|}{67.5/64.9} & \multicolumn{1}{c|}{77.7/76.6} & \multicolumn{1}{c|}{\textbf{79.4}/\textbf{77.8}} & \multicolumn{1}{c|}{89.3/84.3} & \multicolumn{1}{c|}{67.6/65.0} & \multicolumn{1}{c|}{77.7/76.6} & \multicolumn{1}{c|}{\textbf{79.4}/\textbf{77.8}} & \multicolumn{1}{c|}{89.3/84.3} & \multicolumn{1}{c|}{67.6/65.0} & \multicolumn{1}{c|}{77.7/76.6} & \multicolumn{1}{c|}{\textbf{79.4}/\textbf{77.8}} & 89.3/84.4 \\ \hline
\multicolumn{1}{c}{\multirow{2}{*}{DS}}  & C   & \multicolumn{1}{c|}{67.4/74.8} & \multicolumn{1}{c|}{71.6/75.5} & \multicolumn{1}{c|}{\textbf{72.5}/\textbf{75.9}} & \multicolumn{1}{c|}{76.8/77.2} & \multicolumn{1}{c|}{67.1/74.8} & \multicolumn{1}{c|}{71.5/75.4} & \multicolumn{1}{c|}{\textbf{72.4}/\textbf{75.8}} & \multicolumn{1}{c|}{76.8/77.2} & \multicolumn{1}{c|}{66.1/67.3} & \multicolumn{1}{c|}{71.6/\textbf{74.8}} & \multicolumn{1}{c|}{\textbf{71.7}/74.1} & \multicolumn{1}{c|}{85.8/82.0} & \multicolumn{1}{c|}{66.0/67.3} & \multicolumn{1}{c|}{71.6/\textbf{74.9}} & \multicolumn{1}{c|}{\textbf{71.7}/74.1} & \multicolumn{1}{c|}{85.8/82.0} & \multicolumn{1}{c|}{66.1/67.3} & \multicolumn{1}{c|}{71.6/\textbf{74.9}} & \multicolumn{1}{c|}{\textbf{71.7}/74.1} & 85.8/82.0 \\ \cline{2-22} 
\multicolumn{1}{c}{}                     & C++ & \multicolumn{1}{c|}{71.6/75.0} & \multicolumn{1}{c|}{73.8/75.0} & \multicolumn{1}{c|}{\textbf{74.5}/\textbf{75.5}} & \multicolumn{1}{c|}{79.5/75.4} & \multicolumn{1}{c|}{71.2/74.9} & \multicolumn{1}{c|}{73.5/75.0} & \multicolumn{1}{c|}{\textbf{74.1}/\textbf{75.3}} & \multicolumn{1}{c|}{79.6/75.7} & \multicolumn{1}{c|}{67.4/62.8} & \multicolumn{1}{c|}{72.4/73.7} & \multicolumn{1}{c|}{\textbf{74.4}/\textbf{74.9}} & \multicolumn{1}{c|}{88.0/83.1} & \multicolumn{1}{c|}{67.4/62.7} & \multicolumn{1}{c|}{72.4/73.7} & \multicolumn{1}{c|}{\textbf{74.4}/\textbf{74.9}} & \multicolumn{1}{c|}{88.0/83.0} & \multicolumn{1}{c|}{67.4/62.8} & \multicolumn{1}{c|}{72.4/73.7} & \multicolumn{1}{c|}{\textbf{74.4}/\textbf{74.9}} & 88.0/83.1 \\ \hline
\multicolumn{1}{c}{\multirow{2}{*}{C++}} & C   & \multicolumn{1}{c|}{67.4/74.8} & \multicolumn{1}{c|}{74.2/76.0} & \multicolumn{1}{c|}{\textbf{74.7}/\textbf{76.8}} & \multicolumn{1}{c|}{76.8/77.2} & \multicolumn{1}{c|}{67.1/74.8} & \multicolumn{1}{c|}{74.2/76.1} & \multicolumn{1}{c|}{\textbf{74.6}/\textbf{76.6}} & \multicolumn{1}{c|}{76.8/77.2} & \multicolumn{1}{c|}{66.1/67.3} & \multicolumn{1}{c|}{72.8/75.3} & \multicolumn{1}{c|}{\textbf{73.4}/\textbf{75.5}} & \multicolumn{1}{c|}{85.8/82.0} & \multicolumn{1}{c|}{66.0/67.3} & \multicolumn{1}{c|}{72.8/75.3} & \multicolumn{1}{c|}{\textbf{73.4}/\textbf{75.5}} & \multicolumn{1}{c|}{85.8/82.0} & \multicolumn{1}{c|}{66.1/67.3} & \multicolumn{1}{c|}{72.8/75.2} & \multicolumn{1}{c|}{\textbf{73.1}/\textbf{75.3}} & 85.8/82.0 \\ \cline{2-22} 
\multicolumn{1}{c}{}                     & DS  & \multicolumn{1}{c|}{71.8/75.4} & \multicolumn{1}{c|}{71.9/76.5} & \multicolumn{1}{c|}{\textbf{73.3}/\textbf{76.6}} & \multicolumn{1}{c|}{77.2/78.2} & \multicolumn{1}{c|}{71.1/75.2} & \multicolumn{1}{c|}{71.4/76.3} & \multicolumn{1}{c|}{\textbf{72.7}/\textbf{76.4}} & \multicolumn{1}{c|}{77.2/78.2} & \multicolumn{1}{c|}{67.5/64.9} & \multicolumn{1}{c|}{76.4/76.5} & \multicolumn{1}{c|}{\textbf{77.0}/\textbf{76.9}} & \multicolumn{1}{c|}{89.3/84.3} & \multicolumn{1}{c|}{67.6/65.0} & \multicolumn{1}{c|}{76.4/76.5} & \multicolumn{1}{c|}{\textbf{77.0}/\textbf{76.9}} & \multicolumn{1}{c|}{89.3/84.3} & \multicolumn{1}{c|}{67.6/65.0} & \multicolumn{1}{c|}{76.4/76.5} & \multicolumn{1}{c|}{\textbf{77.0}/\textbf{76.9}} & 89.3/84.4 \\ \hline
\multicolumn{2}{c|}{Metrics}                    & \multicolumn{20}{c}{AUC(\%)↑@5/ACC(\%)↑@5}                                                                                                                                                                                                                                                                                                                                                                                                                                                                                                                                                                                                                  \\ \hline
\multicolumn{1}{c}{\multirow{2}{*}{C}}   & C++ & \multicolumn{1}{c|}{74.2/75.6} & \multicolumn{1}{c|}{\textbf{78.9}/\textbf{77.9}} & \multicolumn{1}{c|}{78.8/76.9} & \multicolumn{1}{c|}{79.2/74.7} & \multicolumn{1}{c|}{72.0/75.0} & \multicolumn{1}{c|}{\textbf{78.2}/\textbf{77.3}} & \multicolumn{1}{c|}{77.5/76.1} & \multicolumn{1}{c|}{79.7/76.3} & \multicolumn{1}{c|}{67.5/63.1} & \multicolumn{1}{c|}{74.8/74.6} & \multicolumn{1}{c|}{\textbf{79.3}/\textbf{77.4}} & \multicolumn{1}{c|}{88.0/83.0} & \multicolumn{1}{c|}{67.5/62.9} & \multicolumn{1}{c|}{74.7/74.6} & \multicolumn{1}{c|}{\textbf{79.3}/\textbf{77.4}} & \multicolumn{1}{c|}{88.0/83.1} & \multicolumn{1}{c|}{67.6/63.0} & \multicolumn{1}{c|}{74.7/74.7} & \multicolumn{1}{c|}{\textbf{79.3}/\textbf{77.4}} & 88.0/83.1 \\ \cline{2-22} 
\multicolumn{1}{c}{}                     & DS  & \multicolumn{1}{c|}{74.2/76.0} & \multicolumn{1}{c|}{\textbf{76.9}/\textbf{77.7}} & \multicolumn{1}{c|}{76.6/76.5} & \multicolumn{1}{c|}{77.4/78.3} & \multicolumn{1}{c|}{71.5/75.4} & \multicolumn{1}{c|}{\textbf{75.0}/\textbf{76.9}} & \multicolumn{1}{c|}{75.0/75.7} & \multicolumn{1}{c|}{77.2/78.1} & \multicolumn{1}{c|}{67.9/65.3} & \multicolumn{1}{c|}{77.9/76.7} & \multicolumn{1}{c|}{\textbf{79.5}/\textbf{77.9}} & \multicolumn{1}{c|}{89.2/84.2} & \multicolumn{1}{c|}{67.7/65.1} & \multicolumn{1}{c|}{77.8/76.7} & \multicolumn{1}{c|}{\textbf{79.5}.\textbf{77.9}} & \multicolumn{1}{c|}{89.3/84.4} & \multicolumn{1}{c|}{67.8/65.2} & \multicolumn{1}{c|}{77.8/76.7} & \multicolumn{1}{c|}{\textbf{79.5}/\textbf{77.9}} & 89.3/84.4 \\ \hline
\multicolumn{1}{c}{\multirow{2}{*}{DS}}  & C   & \multicolumn{1}{c|}{68.5/75.0} & \multicolumn{1}{c|}{72.4/75.7} & \multicolumn{1}{c|}{\textbf{73.1}/\textbf{76.1}} & \multicolumn{1}{c|}{76.8/77.3} & \multicolumn{1}{c|}{67.4/74.8} & \multicolumn{1}{c|}{71.8/75.6} & \multicolumn{1}{c|}{\textbf{72.6}/\textbf{75.9}} & \multicolumn{1}{c|}{76.7/77.2} & \multicolumn{1}{c|}{66.3/67.5} & \multicolumn{1}{c|}{71.6/\textbf{74.8}} & \multicolumn{1}{c|}{\textbf{71.8}/74.1} & \multicolumn{1}{c|}{85.9/82.1} & \multicolumn{1}{c|}{66.1/67.4} & \multicolumn{1}{c|}{71.6/\textbf{74.9}} & \multicolumn{1}{c|}{\textbf{71.7}/74.1} & \multicolumn{1}{c|}{85.8/82.0} & \multicolumn{1}{c|}{66.2/67.4} & \multicolumn{1}{c|}{71.7/\textbf{74.9}} & \multicolumn{1}{c|}{\textbf{71.7}/74.1} & 85.8/82.1 \\ \cline{2-22} 
\multicolumn{1}{c}{}                     & C++ & \multicolumn{1}{c|}{74.2/75.6} & \multicolumn{1}{c|}{75.3/75.7} & \multicolumn{1}{c|}{\textbf{76.3}/\textbf{76.4}} & \multicolumn{1}{c|}{79.2/74.7} & \multicolumn{1}{c|}{72.0/75.0} & \multicolumn{1}{c|}{74.0/75.2} & \multicolumn{1}{c|}{\textbf{74.9}/\textbf{75.8}} & \multicolumn{1}{c|}{79.7/76.3} & \multicolumn{1}{c|}{67.5/63.1} & \multicolumn{1}{c|}{72.6/73.8} & \multicolumn{1}{c|}{\textbf{74.5}/\textbf{75.0}} & \multicolumn{1}{c|}{88.0/83.0} & \multicolumn{1}{c|}{67.5/62.9} & \multicolumn{1}{c|}{72.5/73.8} & \multicolumn{1}{c|}{\textbf{74.5}/\textbf{75.0}} & \multicolumn{1}{c|}{88.0/83.1} & \multicolumn{1}{c|}{67.6/63.0} & \multicolumn{1}{c|}{72.5/73.8} & \multicolumn{1}{c|}{\textbf{74.5}/\textbf{75.0}} & 88.0/83.1 \\ \hline
\multicolumn{1}{c}{\multirow{2}{*}{C++}} & C   & \multicolumn{1}{c|}{68.5/75.0} & \multicolumn{1}{c|}{74.4/75.9} & \multicolumn{1}{c|}{\textbf{74.8}/\textbf{76.9}} & \multicolumn{1}{c|}{76.8/77.3} & \multicolumn{1}{c|}{67.4/74.8} & \multicolumn{1}{c|}{74.4/76.2} & \multicolumn{1}{c|}{\textbf{74.7}/\textbf{76.7}} & \multicolumn{1}{c|}{76.7/77.2} & \multicolumn{1}{c|}{66.3/67.5} & \multicolumn{1}{c|}{72.9/75.4} & \multicolumn{1}{c|}{\textbf{73.5}/\textbf{75.5}} & \multicolumn{1}{c|}{85.9/82.1} & \multicolumn{1}{c|}{66.1/67.4} & \multicolumn{1}{c|}{72.9/75.4} & \multicolumn{1}{c|}{\textbf{73.4}/\textbf{75.5}} & \multicolumn{1}{c|}{85.8/82.0} & \multicolumn{1}{c|}{66.2/67.4} & \multicolumn{1}{c|}{72.9/75.3} & \multicolumn{1}{c|}{\textbf{73.2}/\textbf{75.3}} & 85.8/82.1 \\ \cline{2-22} 
\multicolumn{1}{c}{}                     & DS  & \multicolumn{1}{c|}{74.2/76.0} & \multicolumn{1}{c|}{73.4/77.2} & \multicolumn{1}{c|}{\textbf{75.1}/\textbf{77.3}} & \multicolumn{1}{c|}{77.4/78.3} & \multicolumn{1}{c|}{71.5/75.4} & \multicolumn{1}{c|}{71.8/76.4} & \multicolumn{1}{c|}{\textbf{73.1}/\textbf{76.5}} & \multicolumn{1}{c|}{77.2/78.1} & \multicolumn{1}{c|}{67.9/65.3} & \multicolumn{1}{c|}{76.6/76.7} & \multicolumn{1}{c|}{\textbf{77.1}/\textbf{77.0}} & \multicolumn{1}{c|}{89.2/84.2} & \multicolumn{1}{c|}{67.7/65.1} & \multicolumn{1}{c|}{76.5/76.6} & \multicolumn{1}{c|}{\textbf{77.0}/\textbf{77.0}} & \multicolumn{1}{c|}{89.3/84.4} & \multicolumn{1}{c|}{67.8/65.2} & \multicolumn{1}{c|}{76.5/76.6} & \multicolumn{1}{c|}{\textbf{77.0}/\textbf{77.0}} & 89.3/84.4 \\ \hline
\bottomrule
\end{tabular}}
  \caption{The AUC/ACC performance in six scenarios. The best results are highlighted in bold, while $^*$ denotes the upper bounds.}
\label{RQ1}
\end{table*}
\section{Experiments}
In this section, we conduct experiments with the aim of addressing the following questions: 

\begin{itemize}
    \item \textbf{RQ1:}~Can DCSR utilize prior information from a single domain or multiple domains to improve the cold start performance of existing CAT systems?
    \item \textbf{RQ2:}~How effective are the key components of the DCSR framework?
    \item \textbf{RQ3:}~Does DCSR alleviate the issue of the question selection algorithm falling into local optima?
    \item \textbf{RQ4:}~Can DCSR enhance the cold start effect of cognitive diagnosis models?
    \item \textbf{RQ5:}~Is the initial ability assigned by DCSR to cold start examinees reasonable?
\end{itemize}

\subsection{Experimental Settings}
In this section, we introduce the datasets, the selected baselines, and the application of CAT.

\subsubsection{Dataset Description} 
We conducted experiments on five real-world datasets collected from the publicly available PTADisc dataset~\cite{hu2023ptadisc}, covering courses in Data Structures~(DS), C, C++, Python, and Java programming languages. These datasets are sourced from the PTA platform, which records the learning performance of examinees across a series of courses. Each dataset provides the response records of the examinees and question-concept relation, with each dataset considered as a distinct domain. We first excluded examinees with fewer than 100 response records from each dataset. 
The statistics of the processed datasets are shown in Table~\ref{dataset_table}. 
For fairness in testing, we randomly split the filtered examinee records into a training set and a test set in an 80\%:20\% ratio. Examinees in the test set are considered cold-start examinees in the current~(target) domain, meaning their response records in other domains are included in the training set. The training set is only used for training DSCR and the learning-based question selection algorithm, preventing data leakage.

\subsubsection{Baseline Methods}
To demonstrate the effectiveness and compatibility of our framework, we applied it to four widely used CAT systems, including the strategy-based Fisher~\cite{mfi}, MAAT~\cite{maat}, BECAT~\cite{becat}, and the data-driven NCAT~\cite{ncat}. We select cross-domain baselines for comparison, where the Random and Oracle methods represent the lower and upper bounds of CAT cold-start performance, respectively.
\begin{itemize}
    \item Random: This method randomly predicts the initial ability of examinees in the target domain from a $Uniform(0,1)$ distribution, which is the most common method in existing CAT system.
    \item MLCCM: A cross-course method based on meta-learning, applying the idea of meta-learning to learn cross-domain mapping functions from the training set.
    \item Oracle: This method uses CDM to directly train the target domain ability from the response records of examinees in the target domain~(test set).
\end{itemize}

\subsubsection{Evaluation Metrics}
The performance of our DCSR will be validated during the testing phase of the CAT system. We evaluate the accuracy of the final ability estimation by predicting examinees' binary responses to the test question set. For this purpose, we use the area under the Area Under ROC Curve~(AUC)~\cite{auc, yang2021gradient} and Accuracy~(ACC) as evaluation metrics.

\subsubsection{Parameter Settings.} 
In the pre-training phase, for IRT, we set the latent feature dimension of both examinees and questions to 1, while for NCD, it is set to the number of knowledge concepts in the corresponding domain. Additionally, we uniformly set the batch size and learning rate to 32 and 0.002, respectively, for this phase. In the DSCR training phase, the forward process of the diffusion module is set to 1000 steps of noise addition, and DPM-solver~\cite{dpmsolver} is used to accelerate sampling, which is performed in 30 steps. Meanwhile, the batch size and learning rate are fixed at 256 and 0, respectively, in this phase. We initialize all parameters using Xavier~\cite{xavier}, and use the Adam~\cite{adam} optimizer. In the CAT testing phase, the question selection algorithms follow the settings in the original papers, with the test length set to 1 and 5, and the random seed in all the above processes is set to 0. All experiments are conducted on an NVIDIA RTX4090 GPU.

\subsection{Overall Performance~(RQ1)}
\begin{figure}[t]
 \vspace{-5pt}
 \centering
 \includegraphics[width=1\linewidth]{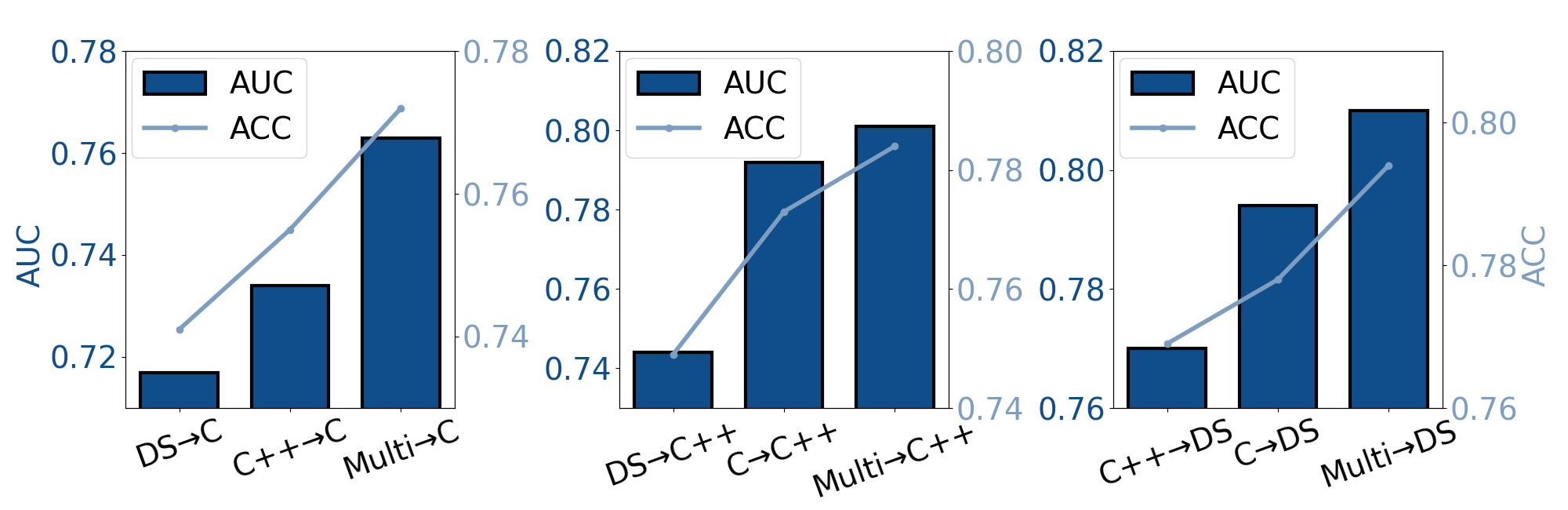}
 \caption{Comparison of the prior information provided by multi and single domain in the NCD-MAAT CAT system.}
 \label{multiPlot}
\end{figure}
To verify the effectiveness of the proposed framework in addressing the CSIP challenge, we compared DCSR with other cross-domain baselines using both single and multi domain as prior knowledge, setting the question selection steps to 1 and 5. First, we explored leveraging a single domain as the source domain, encompassing six cross-domain scenarios. We rotated each dataset to play the role of the target domain, with other datasets serving as the source domain. The experimental results presented in Table~\ref{RQ1} show that our proposed DCSR not only outperforms all baselines in the CAT cold-start task across all six scenarios but also adapts well to 
strategy-based and learning-based selection algorithms. Additionally, we observed the following: (1) Compared to the commonly used Random method, DCSR provides more accurate initial ability estimates for cold-start examinees, especially when there is a significant correlation between the source and target domains, such as C and C++, where DCSR clearly outperforms other baselines and approaches the performance of the Oracle method. This demonstrates the effectiveness of using examinees' domain-shared cognition as the transfer condition. (2) Compared to MLCCM, DCSR does not rely heavily on supervised training, thereby avoiding overfitting to the labels. Moreover, in most scenarios, the ability assigned by DCSR at the initial stage shows better performance than the results diagnosed by other methods after multiple rounds of selection.

Next, we explored using multi-domain information as the source domain, covering three scenarios where C, C++, and DS were set as the target domains, with other courses serving as the source domain. As shown in Figure~\ref{multiPlot}, compared to single-domain information, the multi-domain enriches the common cognition, and the correlation between courses significantly helps resolve the CSIP. The performance in the first two scenarios is particularly effective due to the overlap of concepts between programming languages.
\begin{figure}[t]
 \centering
 \includegraphics[width=1\linewidth]{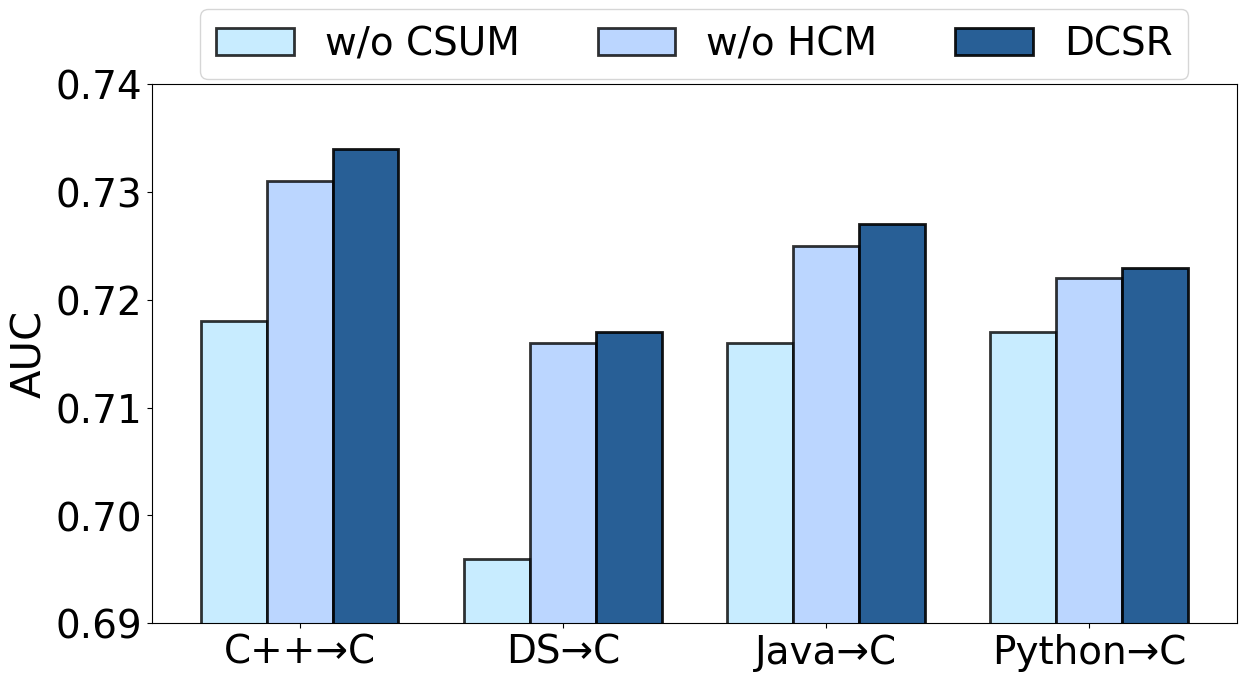}
 \caption{Ablation study on key components.}
 \label{ablation}
\end{figure}
\subsection{Ablation Study~(RQ2)}
This section provides an in-depth analysis of how key components in DCSR contribute to addressing the CSIP challenge. We conducted experiments by individually removing the Cognitive State Unification Module~(CSUM) in section ~\ref{sec:CSUM}~(denoted as w/o CSUM) and the Harmonization and Calibration Module~(HCM) in section~\ref{sec:HCM}~(denoted as w/o HCM). CSUM provides diffusion guidance for the Diffusion Module~(DM), while HCM constrains the DM's output to match the CAT task. As shown in Figure~\ref{ablation}, we validated the impact of different components on DCSR in the NCD-MAAT CAT system, with C set as the target domain and other domains as the source domains. Specifically, when CSUM is removed from DSCR, performance drops significantly, indicating that the specific cognition of the source domain, as a confounding variable, hinders the true causal effect, leading to negative transfer. Similarly, when HCM is removed, there is also a certain degree of performance decline, suggesting that excessive uncertainty affects the alignment of the generated results with the CAT task. Therefore, all components contribute to DCSR to varying degrees.
\subsection{Performance under long test steps~(RQ3)}
\begin{figure}[!t]
 \centering
 \includegraphics[width=0.9\linewidth]{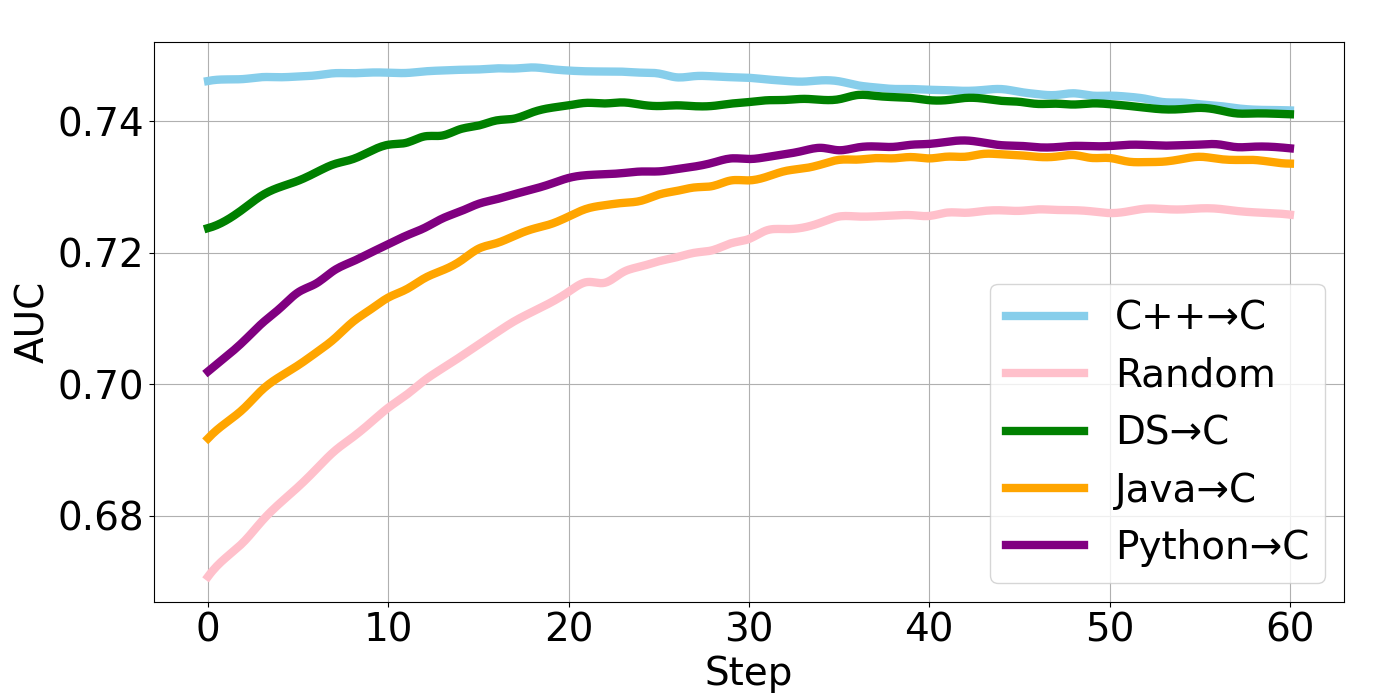}
 \caption{Results under extended testing procedures in the IRT-Fisher CAT system.}
 \label{LongTest}
\end{figure}
To explore the impact of DCSR on cold-start performance under a greedy algorithm, as shown in Figure~\ref{LongTest}, we extended the test length in the IRT-Fisher CAT system and used single-domain information as the prior information provider, with the commonly used Random method as the baseline. We observed that in the early stages of testing, regardless of the source domain, cold-start performance improves significantly. Notably, when C++ is used as the source domain, the initial effect already surpasses the Random method after 60 rounds of question selection due to the correlation between courses. This demonstrates that DCSR not only alleviates the issue of selection algorithms falling into local optima but also guides the selection algorithm by providing an accurate starting point for testing. Additionally, even when a weakly correlated course is used as the source domain, the optimal result of the traditional Random method can be achieved within a shorter test length.
\begin{figure}[tp]
 \centering
 \includegraphics[width=1\linewidth]{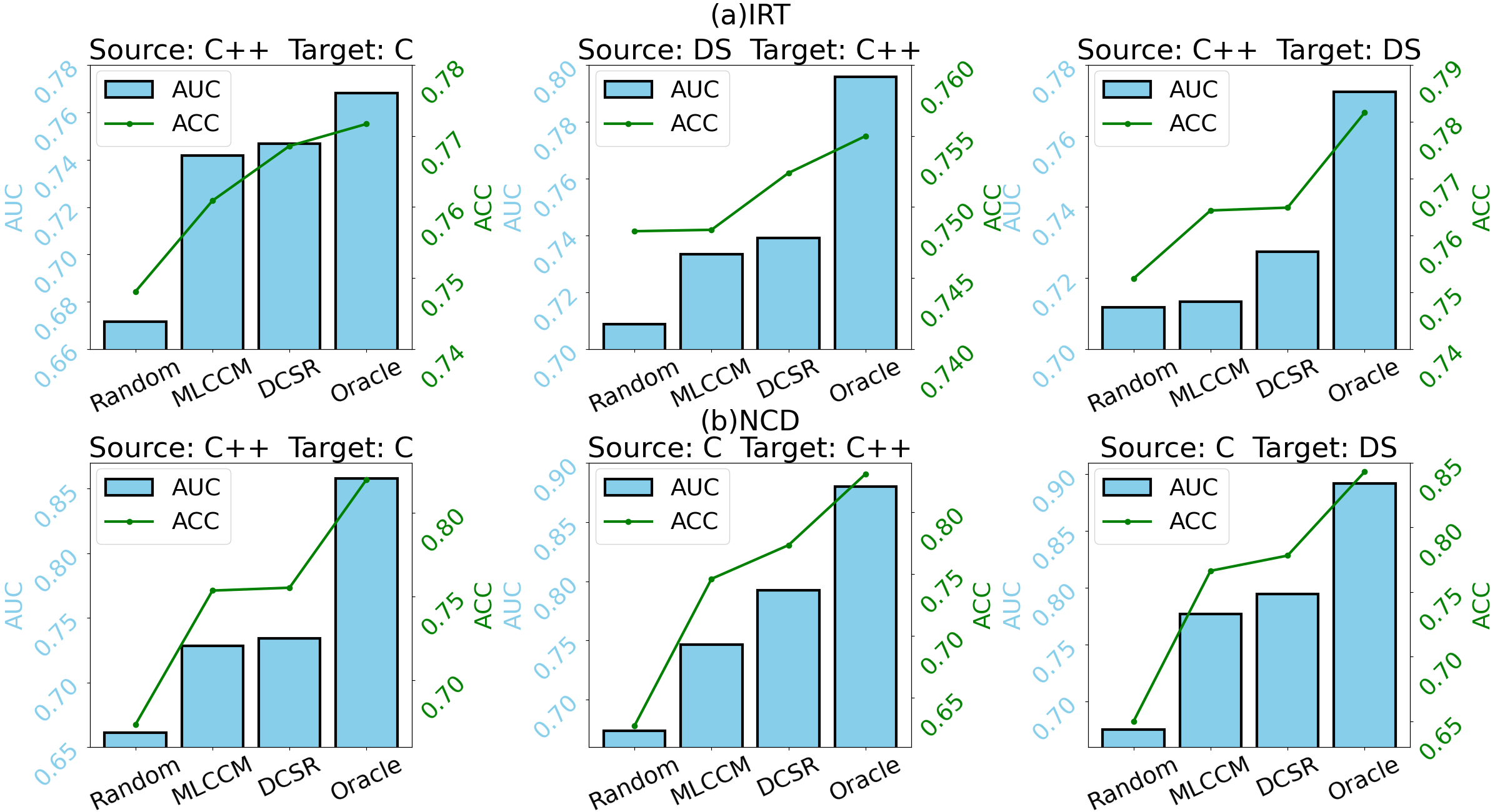}
 \caption{Performance on cognitive diagnosis cold start task.}
 \label{DZCD}
\end{figure}
\subsection{Cold-Start in Cognitive Diagnosis~(RQ4)}
To explore the performance of DCSR in addressing the cold-start problem in cognitive diagnosis task, we directly applied DCSR to CDM, including unidimensional IRT and multidimensional NCD. As shown in Figure~\ref{DZCD}, DCSR demonstrates superior performance in CDM across different dimensions, approaching Oracle performance in some scenarios. Moreover, in high-dimensional CDM, i.e., NCD, DCSR further enhances cold-start performance in CD task. It significantly outperforms the commonly used Random method. Additionally, in scenarios where the correlation is not obvious, such as using DS as the source domain and C++ as the target domain, DCSR shows better performance than MLCCM, significantly outperforming baselines.
\begin{figure}[tp]
 \centering
 \includegraphics[width=1\linewidth]{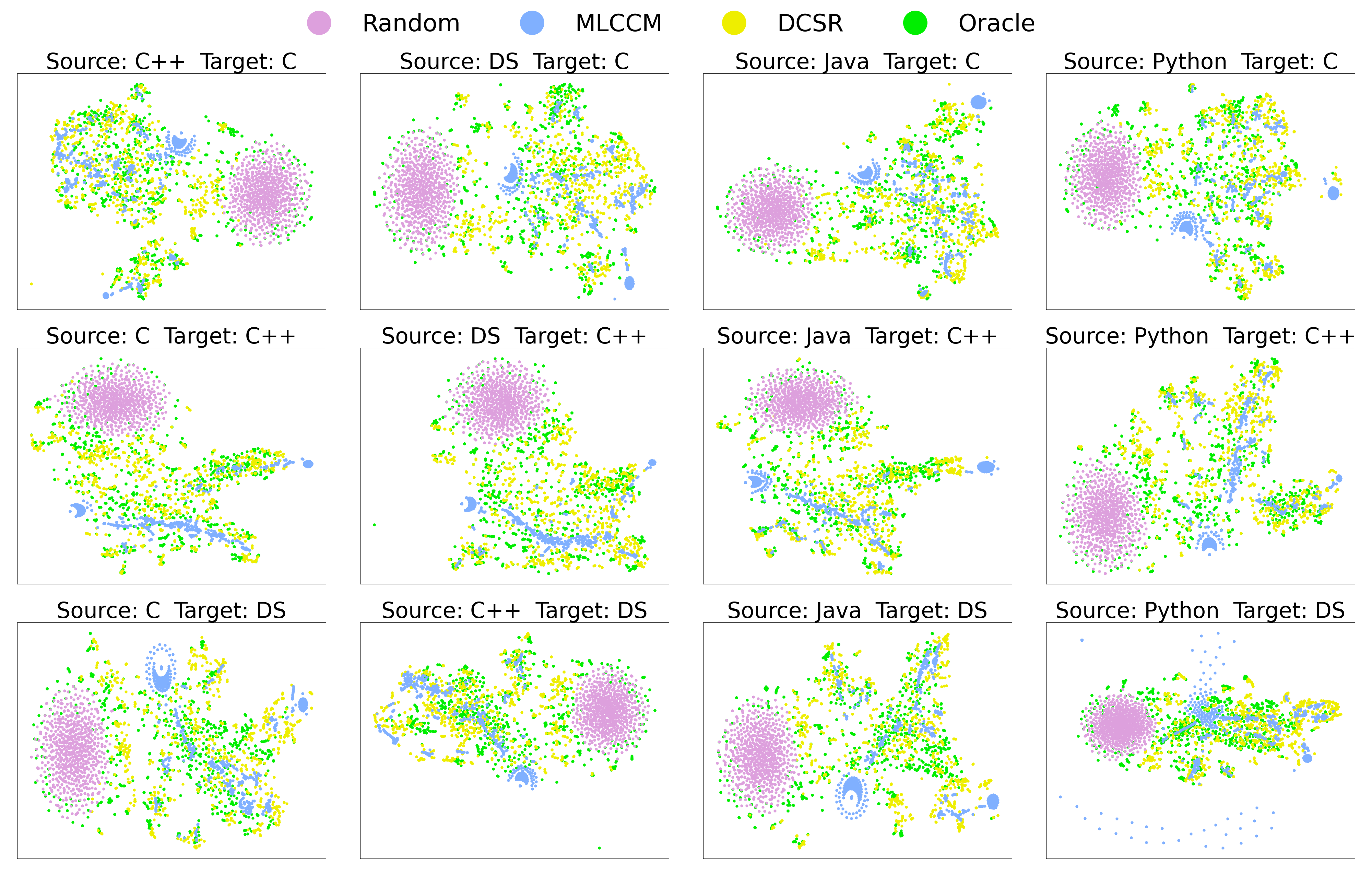}
 \caption{T-SNE visualization of abilities in the target domain.}
 \label{tsne}
\end{figure}
\subsection{Analysis of Initialized Ability~(RQ5)}
\begin{figure}[!tp]
 \centering
 \includegraphics[width=1\linewidth]{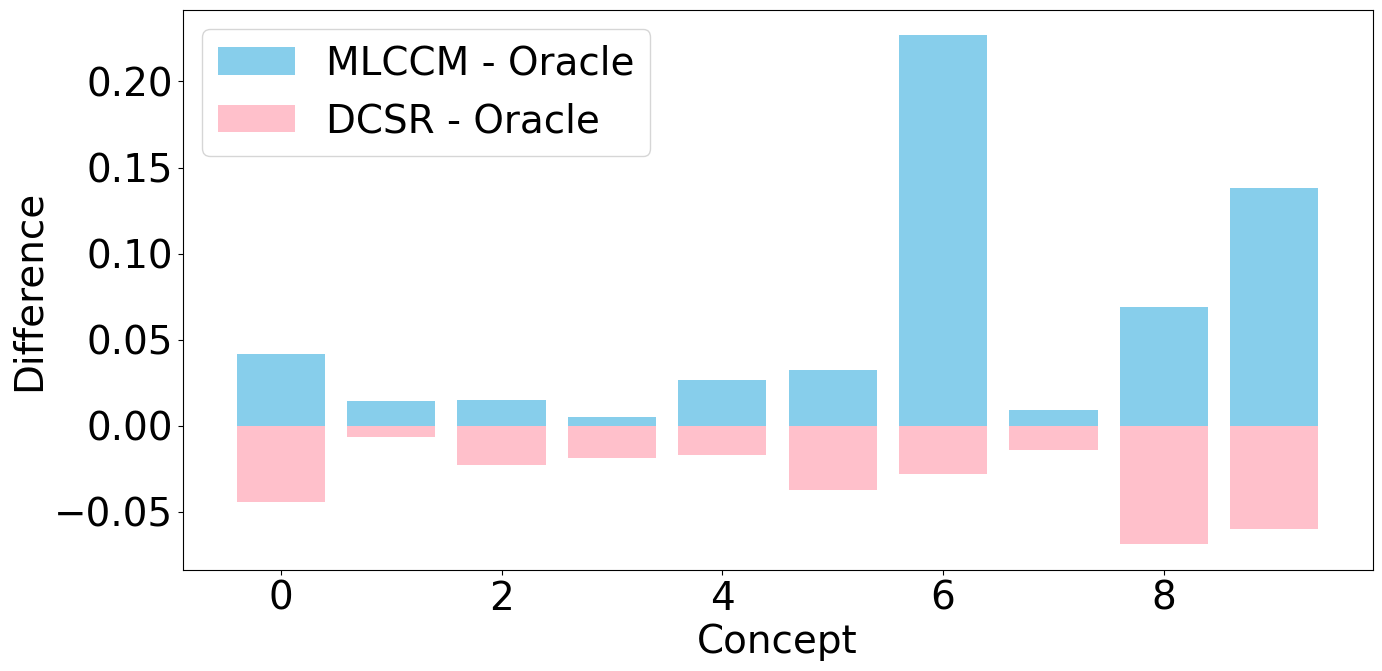}
 \caption{Case analysis of an individual examinee.}
 \label{casestudy}
\end{figure}

We further explored whether the abilities assigned by DCSR to examinees in the target domain are reasonable. We conducted the analysis from two aspects. First, using high-dimensional NCD as the backbone, we visualize the distribution of initial abilities assigned to examinees by different methods across 12 scenarios. As shown in Figure~\ref{tsne}, the ability distribution assigned by the Random method is clearly spherical, which is not conducive to distinguishing between examinees' abilities. In contrast to the localized clustering presented by MLCCM, DCSR is closer to that exhibited by Oracle, showing clear differentiation. This demonstrates that introducing examinees' cross-domain common cognition can generate more personalized initial abilities. Secondly, we randomly selected an examinee, obtained his diagnostic feedback using NCD, and randomly selected ten knowledge concepts to calculate the difference between the diagnosis results of the corresponding dimensions and the results of Oracle. As shown in Figure~\ref{casestudy}, DCSR shows significantly lower fluctuation and tends to underestimate the ability of examinee rather than overestimate him as MLCCM does. This approach better serves the CAT system, as underestimating ability can guide the selection algorithm to choose relatively simpler question, avoiding the negative emotions associated with overly difficult question.
\section{Conclusion}
In this paper, we propose the Diffusion Cognitive States Transfer Framework to address the Cold Start with Insufficient Prior~(CSIP) in Computerized Adaptive Testing~(CAT). This challenge compels CAT systems to use additional selection steps to mitigate the cold start issue. To address this, we first reconstruct the examinee's initial ability in the target domain, guided by diffusion principle. Concurrently, we analyze the causal relationships in the generated outcomes from a model-based causal perspective, proposing three decoupling strategies to block the two backdoor paths that hinder causal discovery. Subsequently, we introduce constraints from the perspectives of consistency and task-oriented to enforce alignment of the generated outcomes with the CAT system. Therefore, this framework can be applied to existing mainstream CAT systems. Finally, extensive experiments highlight the effectiveness and applicability of our framework.

\bibliographystyle{ACM-Reference-Format}
\bibliography{samplebase}


\begin{thebibliography}{50}


\ifx \showCODEN    \undefined \def \showCODEN     #1{\unskip}     \fi
\ifx \showDOI      \undefined \def \showDOI       #1{#1}\fi
\ifx \showISBNx    \undefined \def \showISBNx     #1{\unskip}     \fi
\ifx \showISBNxiii \undefined \def \showISBNxiii  #1{\unskip}     \fi
\ifx \showISSN     \undefined \def \showISSN      #1{\unskip}     \fi
\ifx \showLCCN     \undefined \def \showLCCN      #1{\unskip}     \fi
\ifx \shownote     \undefined \def \shownote      #1{#1}          \fi
\ifx \showarticletitle \undefined \def \showarticletitle #1{#1}   \fi
\ifx \showURL      \undefined \def \showURL       {\relax}        \fi
\providecommand\bibfield[2]{#2}
\providecommand\bibinfo[2]{#2}
\providecommand\natexlab[1]{#1}
\providecommand\showeprint[2][]{arXiv:#2}

\bibitem[Bi et~al\mbox{.}(2020)]%
        {maat}
\bibfield{author}{\bibinfo{person}{Haoyang Bi}, \bibinfo{person}{Haiping Ma}, \bibinfo{person}{Zhenya Huang}, \bibinfo{person}{Yu Yin}, \bibinfo{person}{Qi Liu}, \bibinfo{person}{Enhong Chen}, \bibinfo{person}{Yu Su}, {and} \bibinfo{person}{Shijin Wang}.} \bibinfo{year}{2020}\natexlab{}.
\newblock \showarticletitle{Quality meets diversity: A model-agnostic framework for computerized adaptive testing}. In \bibinfo{booktitle}{\emph{2020 IEEE International Conference on Data Mining (ICDM)}}. IEEE, \bibinfo{pages}{42--51}.
\newblock


\bibitem[Bradley(1997)]%
        {auc}
\bibfield{author}{\bibinfo{person}{Andrew~P Bradley}.} \bibinfo{year}{1997}\natexlab{}.
\newblock \showarticletitle{The use of the area under the ROC curve in the evaluation of machine learning algorithms}.
\newblock \bibinfo{journal}{\emph{Pattern recognition}} \bibinfo{volume}{30}, \bibinfo{number}{7} (\bibinfo{year}{1997}), \bibinfo{pages}{1145--1159}.
\newblock


\bibitem[Chai et~al\mbox{.}(2023)]%
        {diffusion1}
\bibfield{author}{\bibinfo{person}{Shang Chai}, \bibinfo{person}{Liansheng Zhuang}, {and} \bibinfo{person}{Fengying Yan}.} \bibinfo{year}{2023}\natexlab{}.
\newblock \showarticletitle{Layoutdm: Transformer-based diffusion model for layout generation}. In \bibinfo{booktitle}{\emph{Proceedings of the IEEE/CVF Conference on Computer Vision and Pattern Recognition}}. \bibinfo{pages}{18349--18358}.
\newblock


\bibitem[Chang and Ying(1996)]%
        {kli}
\bibfield{author}{\bibinfo{person}{Hua-Hua Chang} {and} \bibinfo{person}{Zhiliang Ying}.} \bibinfo{year}{1996}\natexlab{}.
\newblock \showarticletitle{A global information approach to computerized adaptive testing}.
\newblock \bibinfo{journal}{\emph{Applied Psychological Measurement}} \bibinfo{volume}{20}, \bibinfo{number}{3} (\bibinfo{year}{1996}), \bibinfo{pages}{213--229}.
\newblock


\bibitem[Embretson and Reise(2013)]%
        {irt}
\bibfield{author}{\bibinfo{person}{Susan~E Embretson} {and} \bibinfo{person}{Steven~P Reise}.} \bibinfo{year}{2013}\natexlab{}.
\newblock \bibinfo{booktitle}{\emph{Item response theory}}.
\newblock \bibinfo{publisher}{Psychology Press}.
\newblock


\bibitem[Farajtabar et~al\mbox{.}(2020)]%
        {orth}
\bibfield{author}{\bibinfo{person}{Mehrdad Farajtabar}, \bibinfo{person}{Navid Azizan}, \bibinfo{person}{Alex Mott}, {and} \bibinfo{person}{Ang Li}.} \bibinfo{year}{2020}\natexlab{}.
\newblock \showarticletitle{Orthogonal gradient descent for continual learning}. In \bibinfo{booktitle}{\emph{International Conference on Artificial Intelligence and Statistics}}. PMLR, \bibinfo{pages}{3762--3773}.
\newblock


\bibitem[Gao et~al\mbox{.}(2021)]%
        {rcd}
\bibfield{author}{\bibinfo{person}{Weibo Gao}, \bibinfo{person}{Qi Liu}, \bibinfo{person}{Zhenya Huang}, \bibinfo{person}{Yu Yin}, \bibinfo{person}{Haoyang Bi}, \bibinfo{person}{Mu-Chun Wang}, \bibinfo{person}{Jianhui Ma}, \bibinfo{person}{Shijin Wang}, {and} \bibinfo{person}{Yu Su}.} \bibinfo{year}{2021}\natexlab{}.
\newblock \showarticletitle{RCD: Relation map driven cognitive diagnosis for intelligent education systems}. In \bibinfo{booktitle}{\emph{Proceedings of the 44th international ACM SIGIR conference on research and development in information retrieval}}. \bibinfo{pages}{501--510}.
\newblock


\bibitem[Gao et~al\mbox{.}(2024)]%
        {0123}
\bibfield{author}{\bibinfo{person}{Weibo Gao}, \bibinfo{person}{Qi Liu}, \bibinfo{person}{Hao Wang}, \bibinfo{person}{Linan Yue}, \bibinfo{person}{Haoyang Bi}, \bibinfo{person}{Yin Gu}, \bibinfo{person}{Fangzhou Yao}, \bibinfo{person}{Zheng Zhang}, \bibinfo{person}{Xin Li}, {and} \bibinfo{person}{Yuanjing He}.} \bibinfo{year}{2024}\natexlab{}.
\newblock \showarticletitle{Zero-1-to-3: Domain-Level Zero-Shot Cognitive Diagnosis via One Batch of Early-Bird Students towards Three Diagnostic Objectives}. In \bibinfo{booktitle}{\emph{Proceedings of the AAAI Conference on Artificial Intelligence}}, Vol.~\bibinfo{volume}{38}. \bibinfo{pages}{8417--8426}.
\newblock


\bibitem[Ghosh and Lan(2021)]%
        {bobcat}
\bibfield{author}{\bibinfo{person}{Aritra Ghosh} {and} \bibinfo{person}{Andrew Lan}.} \bibinfo{year}{2021}\natexlab{}.
\newblock \showarticletitle{Bobcat: Bilevel optimization-based computerized adaptive testing}.
\newblock \bibinfo{journal}{\emph{arXiv preprint arXiv:2108.07386}} (\bibinfo{year}{2021}).
\newblock


\bibitem[Glorot and Bengio(2010)]%
        {xavier}
\bibfield{author}{\bibinfo{person}{Xavier Glorot} {and} \bibinfo{person}{Yoshua Bengio}.} \bibinfo{year}{2010}\natexlab{}.
\newblock \showarticletitle{Understanding the difficulty of training deep feedforward neural networks}. In \bibinfo{booktitle}{\emph{Proceedings of the thirteenth international conference on artificial intelligence and statistics}}. JMLR Workshop and Conference Proceedings, \bibinfo{pages}{249--256}.
\newblock


\bibitem[Ho et~al\mbox{.}(2020)]%
        {diffusion}
\bibfield{author}{\bibinfo{person}{Jonathan Ho}, \bibinfo{person}{Ajay Jain}, {and} \bibinfo{person}{Pieter Abbeel}.} \bibinfo{year}{2020}\natexlab{}.
\newblock \showarticletitle{Denoising diffusion probabilistic models}.
\newblock \bibinfo{journal}{\emph{Advances in neural information processing systems}}  \bibinfo{volume}{33} (\bibinfo{year}{2020}), \bibinfo{pages}{6840--6851}.
\newblock


\bibitem[Hong et~al\mbox{.}(2023)]%
        {secat}
\bibfield{author}{\bibinfo{person}{Yuting Hong}, \bibinfo{person}{Shiwei Tong}, \bibinfo{person}{Wei Huang}, \bibinfo{person}{Yan Zhuang}, \bibinfo{person}{Qi Liu}, \bibinfo{person}{Enhong Chen}, \bibinfo{person}{Xin Li}, {and} \bibinfo{person}{Yuanjing He}.} \bibinfo{year}{2023}\natexlab{}.
\newblock \showarticletitle{Search-Efficient Computerized Adaptive Testing}. In \bibinfo{booktitle}{\emph{Proceedings of the 32nd ACM International Conference on Information and Knowledge Management}}. \bibinfo{pages}{773--782}.
\newblock


\bibitem[Hou et~al\mbox{.}(2024)]%
        {cfdiff}
\bibfield{author}{\bibinfo{person}{Yu Hou}, \bibinfo{person}{Jin-Duk Park}, {and} \bibinfo{person}{Won-Yong Shin}.} \bibinfo{year}{2024}\natexlab{}.
\newblock \showarticletitle{Collaborative Filtering Based on Diffusion Models: Unveiling the Potential of High-Order Connectivity}. In \bibinfo{booktitle}{\emph{Proceedings of the 47th International ACM SIGIR Conference on Research and Development in Information Retrieval}}. \bibinfo{pages}{1360--1369}.
\newblock


\bibitem[Hu et~al\mbox{.}(2023)]%
        {hu2023ptadisc}
\bibfield{author}{\bibinfo{person}{Liya Hu}, \bibinfo{person}{Zhiang Dong}, \bibinfo{person}{Jingyuan Chen}, \bibinfo{person}{Guifeng Wang}, \bibinfo{person}{Zhihua Wang}, \bibinfo{person}{Zhou Zhao}, {and} \bibinfo{person}{Fei Wu}.} \bibinfo{year}{2023}\natexlab{}.
\newblock \showarticletitle{PTADisc: a cross-course dataset supporting personalized learning in cold-start scenarios}.
\newblock \bibinfo{journal}{\emph{Advances in Neural Information Processing Systems}}  \bibinfo{volume}{36} (\bibinfo{year}{2023}), \bibinfo{pages}{44976--44996}.
\newblock


\bibitem[Jiang et~al\mbox{.}(2024)]%
        {diffkg}
\bibfield{author}{\bibinfo{person}{Yangqin Jiang}, \bibinfo{person}{Yuhao Yang}, \bibinfo{person}{Lianghao Xia}, {and} \bibinfo{person}{Chao Huang}.} \bibinfo{year}{2024}\natexlab{}.
\newblock \showarticletitle{Diffkg: Knowledge graph diffusion model for recommendation}. In \bibinfo{booktitle}{\emph{Proceedings of the 17th ACM International Conference on Web Search and Data Mining}}. \bibinfo{pages}{313--321}.
\newblock


\bibitem[Kingma and Ba(2014)]%
        {adam}
\bibfield{author}{\bibinfo{person}{Diederik Kingma} {and} \bibinfo{person}{Jimmy Ba}.} \bibinfo{year}{2014}\natexlab{}.
\newblock \showarticletitle{Adam: A Method for Stochastic Optimization}.
\newblock \bibinfo{journal}{\emph{Computer Science}} (\bibinfo{year}{2014}).
\newblock


\bibitem[Li et~al\mbox{.}(2023)]%
        {diffurec}
\bibfield{author}{\bibinfo{person}{Zihao Li}, \bibinfo{person}{Aixin Sun}, {and} \bibinfo{person}{Chenliang Li}.} \bibinfo{year}{2023}\natexlab{}.
\newblock \showarticletitle{Diffurec: A diffusion model for sequential recommendation}.
\newblock \bibinfo{journal}{\emph{ACM Transactions on Information Systems}} \bibinfo{volume}{42}, \bibinfo{number}{3} (\bibinfo{year}{2023}), \bibinfo{pages}{1--28}.
\newblock


\bibitem[Liu et~al\mbox{.}(2023a)]%
        {diffuasr}
\bibfield{author}{\bibinfo{person}{Qidong Liu}, \bibinfo{person}{Fan Yan}, \bibinfo{person}{Xiangyu Zhao}, \bibinfo{person}{Zhaocheng Du}, \bibinfo{person}{Huifeng Guo}, \bibinfo{person}{Ruiming Tang}, {and} \bibinfo{person}{Feng Tian}.} \bibinfo{year}{2023}\natexlab{a}.
\newblock \showarticletitle{Diffusion augmentation for sequential recommendation}. In \bibinfo{booktitle}{\emph{Proceedings of the 32nd ACM International Conference on Information and Knowledge Management}}. \bibinfo{pages}{1576--1586}.
\newblock


\bibitem[Liu et~al\mbox{.}(2024)]%
        {intro2}
\bibfield{author}{\bibinfo{person}{Qi Liu}, \bibinfo{person}{Yan Zhuang}, \bibinfo{person}{Haoyang Bi}, \bibinfo{person}{Zhenya Huang}, \bibinfo{person}{Weizhe Huang}, \bibinfo{person}{Jiatong Li}, \bibinfo{person}{Junhao Yu}, \bibinfo{person}{Zirui Liu}, \bibinfo{person}{Zirui Hu}, \bibinfo{person}{Yuting Hong}, {et~al\mbox{.}}} \bibinfo{year}{2024}\natexlab{}.
\newblock \showarticletitle{Survey of Computerized Adaptive Testing: A Machine Learning Perspective}.
\newblock \bibinfo{journal}{\emph{arXiv preprint arXiv:2404.00712}} (\bibinfo{year}{2024}).
\newblock


\bibitem[Liu et~al\mbox{.}(2023b)]%
        {liu2023homogeneous}
\bibfield{author}{\bibinfo{person}{Shuhuan Liu}, \bibinfo{person}{Xiaoshan Yu}, \bibinfo{person}{Haiping Ma}, \bibinfo{person}{Ziwen Wang}, \bibinfo{person}{Chuan Qin}, {and} \bibinfo{person}{Xingyi Zhang}.} \bibinfo{year}{2023}\natexlab{b}.
\newblock \showarticletitle{Homogeneous Cohort-Aware Group Cognitive Diagnosis: A Multi-grained Modeling Perspective}. In \bibinfo{booktitle}{\emph{Proceedings of the 32nd ACM International Conference on Information and Knowledge Management}}. \bibinfo{pages}{4094--4098}.
\newblock


\bibitem[Lord(2012)]%
        {mfi}
\bibfield{author}{\bibinfo{person}{Frederic~M Lord}.} \bibinfo{year}{2012}\natexlab{}.
\newblock \bibinfo{booktitle}{\emph{Applications of item response theory to practical testing problems}}.
\newblock \bibinfo{publisher}{Routledge}.
\newblock


\bibitem[Lu et~al\mbox{.}(2022)]%
        {dpmsolver}
\bibfield{author}{\bibinfo{person}{Cheng Lu}, \bibinfo{person}{Yuhao Zhou}, \bibinfo{person}{Fan Bao}, \bibinfo{person}{Jianfei Chen}, \bibinfo{person}{Chongxuan Li}, {and} \bibinfo{person}{Jun Zhu}.} \bibinfo{year}{2022}\natexlab{}.
\newblock \showarticletitle{Dpm-solver: A fast ode solver for diffusion probabilistic model sampling in around 10 steps}.
\newblock \bibinfo{journal}{\emph{Advances in Neural Information Processing Systems}}  \bibinfo{volume}{35} (\bibinfo{year}{2022}), \bibinfo{pages}{5775--5787}.
\newblock


\bibitem[Ma et~al\mbox{.}(2024a)]%
        {ma2024dgcd}
\bibfield{author}{\bibinfo{person}{Haiping Ma}, \bibinfo{person}{Siyu Song}, \bibinfo{person}{Chuan Qin}, \bibinfo{person}{Xiaoshan Yu}, \bibinfo{person}{Limiao Zhang}, \bibinfo{person}{Xingyi Zhang}, {and} \bibinfo{person}{Hengshu Zhu}.} \bibinfo{year}{2024}\natexlab{a}.
\newblock \showarticletitle{DGCD: An Adaptive Denoising GNN for Group-level Cognitive Diagnosis}. In \bibinfo{booktitle}{\emph{The 33rd International Joint Conference on Artificial Intelligence (IJCAI-24)}}.
\newblock


\bibitem[Ma et~al\mbox{.}(2024b)]%
        {cqw}
\bibfield{author}{\bibinfo{person}{Haiping Ma}, \bibinfo{person}{Changqian Wang}, \bibinfo{person}{Hengshu Zhu}, \bibinfo{person}{Shangshang Yang}, \bibinfo{person}{Xiaoming Zhang}, {and} \bibinfo{person}{Xingyi Zhang}.} \bibinfo{year}{2024}\natexlab{b}.
\newblock \showarticletitle{Enhancing cognitive diagnosis using un-interacted exercises: A collaboration-aware mixed sampling approach}. In \bibinfo{booktitle}{\emph{Proceedings of the AAAI Conference on Artificial Intelligence}}, Vol.~\bibinfo{volume}{38}. \bibinfo{pages}{8877--8885}.
\newblock


\bibitem[Ma et~al\mbox{.}(2024c)]%
        {plugdiff}
\bibfield{author}{\bibinfo{person}{Haokai Ma}, \bibinfo{person}{Ruobing Xie}, \bibinfo{person}{Lei Meng}, \bibinfo{person}{Xin Chen}, \bibinfo{person}{Xu Zhang}, \bibinfo{person}{Leyu Lin}, {and} \bibinfo{person}{Zhanhui Kang}.} \bibinfo{year}{2024}\natexlab{c}.
\newblock \showarticletitle{Plug-in diffusion model for sequential recommendation}. In \bibinfo{booktitle}{\emph{Proceedings of the AAAI Conference on Artificial Intelligence}}, Vol.~\bibinfo{volume}{38}. \bibinfo{pages}{8886--8894}.
\newblock


\bibitem[Ma et~al\mbox{.}(2024d)]%
        {ma2024hd}
\bibfield{author}{\bibinfo{person}{Haiping Ma}, \bibinfo{person}{Yong Yang}, \bibinfo{person}{Chuan Qin}, \bibinfo{person}{Xiaoshan Yu}, \bibinfo{person}{Shangshang Yang}, \bibinfo{person}{Xingyi Zhang}, {and} \bibinfo{person}{Hengshu Zhu}.} \bibinfo{year}{2024}\natexlab{d}.
\newblock \showarticletitle{HD-KT: Advancing Robust Knowledge Tracing via Anomalous Learning Interaction Detection}. In \bibinfo{booktitle}{\emph{Proceedings of the ACM on Web Conference 2024}}. \bibinfo{pages}{4479--4488}.
\newblock


\bibitem[Ma et~al\mbox{.}(2023)]%
        {dlcat}
\bibfield{author}{\bibinfo{person}{Haiping Ma}, \bibinfo{person}{Yi Zeng}, \bibinfo{person}{Shangshang Yang}, \bibinfo{person}{Chuan Qin}, \bibinfo{person}{Xingyi Zhang}, {and} \bibinfo{person}{Limiao Zhang}.} \bibinfo{year}{2023}\natexlab{}.
\newblock \showarticletitle{A novel computerized adaptive testing framework with decoupled learning selector}.
\newblock \bibinfo{journal}{\emph{Complex \& Intelligent Systems}} \bibinfo{volume}{9}, \bibinfo{number}{5} (\bibinfo{year}{2023}), \bibinfo{pages}{5555--5566}.
\newblock


\bibitem[Van~der Linden and Glas(2000)]%
        {cattheory}
\bibfield{author}{\bibinfo{person}{Wim~J Van~der Linden} {and} \bibinfo{person}{Cees~AW Glas}.} \bibinfo{year}{2000}\natexlab{}.
\newblock \bibinfo{booktitle}{\emph{Computerized adaptive testing: Theory and practice}}.
\newblock \bibinfo{publisher}{Springer}.
\newblock


\bibitem[Vie et~al\mbox{.}(2017)]%
        {intro3}
\bibfield{author}{\bibinfo{person}{Jill-J{\^e}nn Vie}, \bibinfo{person}{Fabrice Popineau}, \bibinfo{person}{{\'E}ric Bruillard}, {and} \bibinfo{person}{Yolaine Bourda}.} \bibinfo{year}{2017}\natexlab{}.
\newblock \showarticletitle{A review of recent advances in adaptive assessment}.
\newblock \bibinfo{journal}{\emph{Learning analytics: Fundaments, applications, and trends: A view of the current state of the art to enhance e-learning}} (\bibinfo{year}{2017}), \bibinfo{pages}{113--142}.
\newblock


\bibitem[Wainer et~al\mbox{.}(2000)]%
        {intro1}
\bibfield{author}{\bibinfo{person}{Howard Wainer}, \bibinfo{person}{Neil~J Dorans}, \bibinfo{person}{Ronald Flaugher}, \bibinfo{person}{Bert~F Green}, {and} \bibinfo{person}{Robert~J Mislevy}.} \bibinfo{year}{2000}\natexlab{}.
\newblock \bibinfo{booktitle}{\emph{Computerized adaptive testing: A primer}}.
\newblock \bibinfo{publisher}{Routledge}.
\newblock


\bibitem[Walker et~al\mbox{.}(2022)]%
        {codigem}
\bibfield{author}{\bibinfo{person}{Joojo Walker}, \bibinfo{person}{Ting Zhong}, \bibinfo{person}{Fengli Zhang}, \bibinfo{person}{Qiang Gao}, {and} \bibinfo{person}{Fan Zhou}.} \bibinfo{year}{2022}\natexlab{}.
\newblock \showarticletitle{Recommendation via collaborative diffusion generative model}. In \bibinfo{booktitle}{\emph{International Conference on Knowledge Science, Engineering and Management}}. Springer, \bibinfo{pages}{593--605}.
\newblock


\bibitem[Wang et~al\mbox{.}(2020)]%
        {ncd}
\bibfield{author}{\bibinfo{person}{Fei Wang}, \bibinfo{person}{Qi Liu}, \bibinfo{person}{Enhong Chen}, \bibinfo{person}{Zhenya Huang}, \bibinfo{person}{Yuying Chen}, \bibinfo{person}{Yu Yin}, \bibinfo{person}{Zai Huang}, {and} \bibinfo{person}{Shijin Wang}.} \bibinfo{year}{2020}\natexlab{}.
\newblock \showarticletitle{Neural cognitive diagnosis for intelligent education systems}. In \bibinfo{booktitle}{\emph{Proceedings of the AAAI conference on artificial intelligence}}, Vol.~\bibinfo{volume}{34}. \bibinfo{pages}{6153--6161}.
\newblock


\bibitem[Wang et~al\mbox{.}(2022)]%
        {kancd}
\bibfield{author}{\bibinfo{person}{Fei Wang}, \bibinfo{person}{Qi Liu}, \bibinfo{person}{Enhong Chen}, \bibinfo{person}{Zhenya Huang}, \bibinfo{person}{Yu Yin}, \bibinfo{person}{Shijin Wang}, {and} \bibinfo{person}{Yu Su}.} \bibinfo{year}{2022}\natexlab{}.
\newblock \showarticletitle{NeuralCD: a general framework for cognitive diagnosis}.
\newblock \bibinfo{journal}{\emph{IEEE Transactions on Knowledge and Data Engineering}} \bibinfo{volume}{35}, \bibinfo{number}{8} (\bibinfo{year}{2022}), \bibinfo{pages}{8312--8327}.
\newblock


\bibitem[Wang et~al\mbox{.}(2023a)]%
        {gmocat}
\bibfield{author}{\bibinfo{person}{Hangyu Wang}, \bibinfo{person}{Ting Long}, \bibinfo{person}{Liang Yin}, \bibinfo{person}{Weinan Zhang}, \bibinfo{person}{Wei Xia}, \bibinfo{person}{Qichen Hong}, \bibinfo{person}{Dingyin Xia}, \bibinfo{person}{Ruiming Tang}, {and} \bibinfo{person}{Yong Yu}.} \bibinfo{year}{2023}\natexlab{a}.
\newblock \showarticletitle{GMOCAT: A Graph-Enhanced Multi-Objective Method for Computerized Adaptive Testing}. In \bibinfo{booktitle}{\emph{Proceedings of the 29th ACM SIGKDD Conference on Knowledge Discovery and Data Mining}}. \bibinfo{pages}{2279--2289}.
\newblock


\bibitem[Wang et~al\mbox{.}(2023b)]%
        {diffrec}
\bibfield{author}{\bibinfo{person}{Wenjie Wang}, \bibinfo{person}{Yiyan Xu}, \bibinfo{person}{Fuli Feng}, \bibinfo{person}{Xinyu Lin}, \bibinfo{person}{Xiangnan He}, {and} \bibinfo{person}{Tat-Seng Chua}.} \bibinfo{year}{2023}\natexlab{b}.
\newblock \showarticletitle{Diffusion recommender model}. In \bibinfo{booktitle}{\emph{Proceedings of the 46th International ACM SIGIR Conference on Research and Development in Information Retrieval}}. \bibinfo{pages}{832--841}.
\newblock


\bibitem[Xuan(2024)]%
        {diffcdr}
\bibfield{author}{\bibinfo{person}{Yuner Xuan}.} \bibinfo{year}{2024}\natexlab{}.
\newblock \showarticletitle{Diffusion Cross-domain Recommendation}.
\newblock \bibinfo{journal}{\emph{arXiv preprint arXiv:2402.02182}} (\bibinfo{year}{2024}).
\newblock


\bibitem[Yang et~al\mbox{.}(2024a)]%
        {yang2024disengcd}
\bibfield{author}{\bibinfo{person}{Shangshang Yang}, \bibinfo{person}{Mingyang Chen}, \bibinfo{person}{Ziwen Wang}, \bibinfo{person}{Xiaoshan Yu}, \bibinfo{person}{Panpan Zhang}, \bibinfo{person}{Haiping Ma}, {and} \bibinfo{person}{Xingyi Zhang}.} \bibinfo{year}{2024}\natexlab{a}.
\newblock \showarticletitle{DisenGCD: A Meta Multigraph-assisted Disentangled Graph Learning Framework for Cognitive Diagnosis}.
\newblock \bibinfo{journal}{\emph{arXiv preprint arXiv:2410.17564}} (\bibinfo{year}{2024}).
\newblock


\bibitem[Yang et~al\mbox{.}(2024b)]%
        {yang2024evolutionary1}
\bibfield{author}{\bibinfo{person}{Shangshang Yang}, \bibinfo{person}{Haiping Ma}, \bibinfo{person}{Ying Bi}, \bibinfo{person}{Ye Tian}, \bibinfo{person}{Limiao Zhang}, \bibinfo{person}{Yaochu Jin}, {and} \bibinfo{person}{Xingyi Zhang}.} \bibinfo{year}{2024}\natexlab{b}.
\newblock \showarticletitle{An evolutionary multi-objective neural architecture search approach to advancing cognitive diagnosis in intelligent education}.
\newblock \bibinfo{journal}{\emph{IEEE Transactions on Evolutionary Computation}} (\bibinfo{year}{2024}).
\newblock


\bibitem[Yang et~al\mbox{.}(2024c)]%
        {yang2024endowing}
\bibfield{author}{\bibinfo{person}{Shangshang Yang}, \bibinfo{person}{Linrui Qin}, {and} \bibinfo{person}{Xiaoshan Yu}.} \bibinfo{year}{2024}\natexlab{c}.
\newblock \showarticletitle{Endowing Interpretability for Neural Cognitive Diagnosis by Efficient Kolmogorov-Arnold Networks}.
\newblock \bibinfo{journal}{\emph{arXiv preprint arXiv:2405.14399}} (\bibinfo{year}{2024}).
\newblock


\bibitem[Yang et~al\mbox{.}(2021)]%
        {yang2021gradient}
\bibfield{author}{\bibinfo{person}{Shangshang Yang}, \bibinfo{person}{Ye Tian}, \bibinfo{person}{Cheng He}, \bibinfo{person}{Xingyi Zhang}, \bibinfo{person}{Kay~Chen Tan}, {and} \bibinfo{person}{Yaochu Jin}.} \bibinfo{year}{2021}\natexlab{}.
\newblock \showarticletitle{A gradient-guided evolutionary approach to training deep neural networks}.
\newblock \bibinfo{journal}{\emph{IEEE Transactions on Neural Networks and Learning Systems}} \bibinfo{volume}{33}, \bibinfo{number}{9} (\bibinfo{year}{2021}), \bibinfo{pages}{4861--4875}.
\newblock


\bibitem[Yang et~al\mbox{.}(2023a)]%
        {yang2023cognitive}
\bibfield{author}{\bibinfo{person}{Shangshang Yang}, \bibinfo{person}{Haoyu Wei}, \bibinfo{person}{Haiping Ma}, \bibinfo{person}{Ye Tian}, \bibinfo{person}{Xingyi Zhang}, \bibinfo{person}{Yunbo Cao}, {and} \bibinfo{person}{Yaochu Jin}.} \bibinfo{year}{2023}\natexlab{a}.
\newblock \showarticletitle{Cognitive diagnosis-based personalized exercise group assembly via a multi-objective evolutionary algorithm}.
\newblock \bibinfo{journal}{\emph{IEEE Transactions on Emerging Topics in Computational Intelligence}} \bibinfo{volume}{7}, \bibinfo{number}{3} (\bibinfo{year}{2023}), \bibinfo{pages}{829--844}.
\newblock


\bibitem[Yang et~al\mbox{.}(2024e)]%
        {yang2024evolutionary}
\bibfield{author}{\bibinfo{person}{Shangshang Yang}, \bibinfo{person}{Xiaoshan Yu}, \bibinfo{person}{Ye Tian}, \bibinfo{person}{Xueming Yan}, \bibinfo{person}{Haiping Ma}, {and} \bibinfo{person}{Xingyi Zhang}.} \bibinfo{year}{2024}\natexlab{e}.
\newblock \showarticletitle{Evolutionary neural architecture search for transformer in knowledge tracing}.
\newblock \bibinfo{journal}{\emph{Advances in Neural Information Processing Systems}}  \bibinfo{volume}{36} (\bibinfo{year}{2024}).
\newblock


\bibitem[Yang et~al\mbox{.}(2023b)]%
        {yang2023evolutionary}
\bibfield{author}{\bibinfo{person}{Shangshang Yang}, \bibinfo{person}{Cheng Zhen}, \bibinfo{person}{Ye Tian}, \bibinfo{person}{Haiping Ma}, \bibinfo{person}{Yuanchao Liu}, \bibinfo{person}{Panpan Zhang}, {and} \bibinfo{person}{Xingyi Zhang}.} \bibinfo{year}{2023}\natexlab{b}.
\newblock \showarticletitle{Evolutionary multi-objective neural architecture search for generalized cognitive diagnosis models}. In \bibinfo{booktitle}{\emph{2023 5th International Conference on Data-driven Optimization of Complex Systems (DOCS)}}. IEEE, \bibinfo{pages}{1--10}.
\newblock


\bibitem[Yang et~al\mbox{.}(2024d)]%
        {dreamrec}
\bibfield{author}{\bibinfo{person}{Zhengyi Yang}, \bibinfo{person}{Jiancan Wu}, \bibinfo{person}{Zhicai Wang}, \bibinfo{person}{Xiang Wang}, \bibinfo{person}{Yancheng Yuan}, {and} \bibinfo{person}{Xiangnan He}.} \bibinfo{year}{2024}\natexlab{d}.
\newblock \showarticletitle{Generate what you prefer: Reshaping sequential recommendation via guided diffusion}.
\newblock \bibinfo{journal}{\emph{Advances in Neural Information Processing Systems}}  \bibinfo{volume}{36} (\bibinfo{year}{2024}).
\newblock


\bibitem[Yu et~al\mbox{.}(2024a)]%
        {yu2024rdgt}
\bibfield{author}{\bibinfo{person}{Xiaoshan Yu}, \bibinfo{person}{Chuan Qin}, \bibinfo{person}{Dazhong Shen}, \bibinfo{person}{Haiping Ma}, \bibinfo{person}{Le Zhang}, \bibinfo{person}{Xingyi Zhang}, \bibinfo{person}{Hengshu Zhu}, {and} \bibinfo{person}{Hui Xiong}.} \bibinfo{year}{2024}\natexlab{a}.
\newblock \showarticletitle{Rdgt: enhancing group cognitive diagnosis with relation-guided dual-side graph transformer}.
\newblock \bibinfo{journal}{\emph{IEEE Transactions on Knowledge and Data Engineering}} (\bibinfo{year}{2024}).
\newblock


\bibitem[Yu et~al\mbox{.}(2024b)]%
        {yu2024rigl}
\bibfield{author}{\bibinfo{person}{Xiaoshan Yu}, \bibinfo{person}{Chuan Qin}, \bibinfo{person}{Dazhong Shen}, \bibinfo{person}{Shangshang Yang}, \bibinfo{person}{Haiping Ma}, \bibinfo{person}{Hengshu Zhu}, {and} \bibinfo{person}{Xingyi Zhang}.} \bibinfo{year}{2024}\natexlab{b}.
\newblock \showarticletitle{Rigl: A unified reciprocal approach for tracing the independent and group learning processes}. In \bibinfo{booktitle}{\emph{Proceedings of the 30th ACM SIGKDD Conference on Knowledge Discovery and Data Mining}}. \bibinfo{pages}{4047--4058}.
\newblock


\bibitem[Yu et~al\mbox{.}(2024c)]%
        {yu2024disco}
\bibfield{author}{\bibinfo{person}{Xiaoshan Yu}, \bibinfo{person}{Chuan Qin}, \bibinfo{person}{Qi Zhang}, \bibinfo{person}{Chen Zhu}, \bibinfo{person}{Haiping Ma}, \bibinfo{person}{Xingyi Zhang}, {and} \bibinfo{person}{Hengshu Zhu}.} \bibinfo{year}{2024}\natexlab{c}.
\newblock \showarticletitle{DISCO: A Hierarchical Disentangled Cognitive Diagnosis Framework for Interpretable Job Recommendation}.
\newblock \bibinfo{journal}{\emph{arXiv preprint arXiv:2410.07671}} (\bibinfo{year}{2024}).
\newblock


\bibitem[Zhao et~al\mbox{.}(2024)]%
        {denoisediff}
\bibfield{author}{\bibinfo{person}{Jujia Zhao}, \bibinfo{person}{Wang Wenjie}, \bibinfo{person}{Yiyan Xu}, \bibinfo{person}{Teng Sun}, \bibinfo{person}{Fuli Feng}, {and} \bibinfo{person}{Tat-Seng Chua}.} \bibinfo{year}{2024}\natexlab{}.
\newblock \showarticletitle{Denoising diffusion recommender model}. In \bibinfo{booktitle}{\emph{Proceedings of the 47th International ACM SIGIR Conference on Research and Development in Information Retrieval}}. \bibinfo{pages}{1370--1379}.
\newblock


\bibitem[Zhuang et~al\mbox{.}(2022)]%
        {ncat}
\bibfield{author}{\bibinfo{person}{Yan Zhuang}, \bibinfo{person}{Qi Liu}, \bibinfo{person}{Zhenya Huang}, \bibinfo{person}{Zhi Li}, \bibinfo{person}{Shuanghong Shen}, {and} \bibinfo{person}{Haiping Ma}.} \bibinfo{year}{2022}\natexlab{}.
\newblock \showarticletitle{Fully adaptive framework: Neural computerized adaptive testing for online education}. In \bibinfo{booktitle}{\emph{Proceedings of the AAAI conference on artificial intelligence}}, Vol.~\bibinfo{volume}{36}. \bibinfo{pages}{4734--4742}.
\newblock


\bibitem[Zhuang et~al\mbox{.}(2024)]%
        {becat}
\bibfield{author}{\bibinfo{person}{Yan Zhuang}, \bibinfo{person}{Qi Liu}, \bibinfo{person}{GuanHao Zhao}, \bibinfo{person}{Zhenya Huang}, \bibinfo{person}{Weizhe Huang}, \bibinfo{person}{Zachary Pardos}, \bibinfo{person}{Enhong Chen}, \bibinfo{person}{Jinze Wu}, {and} \bibinfo{person}{Xin Li}.} \bibinfo{year}{2024}\natexlab{}.
\newblock \showarticletitle{A bounded ability estimation for computerized adaptive testing}.
\newblock \bibinfo{journal}{\emph{Advances in Neural Information Processing Systems}}  \bibinfo{volume}{36} (\bibinfo{year}{2024}).
\newblock


\end{thebibliography}



\end{document}